\documentclass[lettersize,journal]{IEEEtran}
\usepackage{amsmath,amsfonts}
\usepackage{algorithmic}
\usepackage{algorithm}
\usepackage{array}
\usepackage[caption=false,font=normalsize,labelfont=sf,textfont=sf]{subfig}
\usepackage{textcomp}
\usepackage{stfloats}
\usepackage{url}
\usepackage{verbatim}
\usepackage{graphicx}
\usepackage[table]{xcolor}
\usepackage{cite}
\usepackage{booktabs}
\usepackage{multirow}
\begin{document}

\title{ProxyImg: Towards Highly-Controllable Image Representation via Hierarchical Disentangled Proxy Embedding }

\author{Ye Chen$^{\ast}$, Yupeng Zhu$^{\ast}$, Xiongzhen Zhang, Zhewen Wan, Yingzhe Li, Wenjun Zhang, Bingbing Ni$^{\dag}$
\thanks{Ye Chen, Yepeng Zhu, Xiongzhen Zhang, Zhenwen Wan, Wenjun Zhang and Bingbing Ni are with Shanghai Jiao Tong University, Shanghai 200240, China (e-mail: chenye123@sjtu.edu.cn; zhuyupeng@sjtu.edu.cn; nibingbing@sjtu.edu.cn;).}
\thanks{Yingzhe Li is with Huawei Cloud BU (yingzhe.li@huawei.com).}
\thanks{$^{\ast}$Authors contributed equally to this work.}
\thanks{$^{\dag}$Corresponding author: Bingbing Ni.}}



\maketitle

\begin{abstract}
Prevailing image representation methods, whether explicit such as raster images/Gaussian primitives or implicit such as latent images, either suffer from representation redundancy resulting manual editing burden, or lack of direct mapping from latent variables to semantic instances/parts that makes it impossible for fine-grained manipulation, all of which hinder the application of efficient and controllable image/video editing.
To overcome these limitations, we introduce a hierarchical proxy-based parametric image representation that effectively disentangles semantic, geometric, and textural attributes into independent and manipulable parameter spaces. Building upon a semantic-aware decomposition of the input, our novel representation constructs hierarchical proxy geometries through adaptive Bézier fitting with iterative internal region division/meshing. It further embeds multi-scale implicit texture parameters/codes into the above geometric-aware distributed proxy nodes, which collectively enable continuous high-fidelity pixel domain reconstruction (up to arbitrary accuracy) and instance/part-independent semantic editing. Additionally, a locality-adaptive feature/codes indexing mechanism is developed to ensure spatial texture coherence, facilitating high-quality background completion without relying on generative models.
Extensive experiments on image reconstruction and editing benchmarks—including ImageNet, OIR-Bench, and HumanEdit—demonstrate that our method achieves state-of-the-art rendering fidelity with significantly fewer parameters, while enabling intuitive, interactive, and physically plausible manipulation. Moreover, by integrating proxy nodes with Position-Based Dynamics, our framework supports real-time, physics-driven animation using lightweight implicit rendering, achieving superior temporal consistency and visual realism compared with generative approaches.
\end{abstract}

\begin{IEEEkeywords}
Layer-wise Image Representation, Distributed Embedding, Controllable Image Editing, Image Animation.
\end{IEEEkeywords}

\section{Introduction}
\IEEEPARstart{H}{ow} to represent images is a fundamental problem in the field of computer vision. Traditionally, images are represented by pixels stored on fixed and discrete grids, known as \textbf{\textit{Raster Images}}. One main advantage of such representation is that, there is virtually no limit to its ability to express image details with enough pixels. Nevertheless, pixel-based representation is plagued by numerous limitations.
On one hand, it suffers from excessive redundancy due to its storage on fixed grids, which inherently constrains image resolution, along with inevitable information loss during image re-scaling. On the other hand, most importantly, pixel-based representation stores all the information of an image in the RGB values of the pixels, resulting in a high coupling between the representation of image geometry and texture. Hence, it is challenging to edit the shape/geometry and texture of a raster image separately. 

\textbf{\textit{Parametric Images}} embody an alternative paradigm for representing visual information, which models visual contents as a set of parameters. Parametric images can be further categorized into \textbf{\textit{Explicit Parametric Images}} and \textbf{\textit{Implicit Parametric Images}}. A typical explicit parametric representation is the vector image, which encodes visual information using geometric primitives (\emph{e.g.}, curves, polygons) defined by control points in continuous space. Vector images provide three key advantages over raster images: (1) \textbf{Resolution-Agnostic}: continuous parameterization allows arbitrary-resolution rendering without rescaling artifacts; (2) \textbf{Easy Editing}: explicit parameters enable structured shape and color modifications; (3) \textbf{Compactness}: mathematical primitives require far fewer parameters than raster grids. However, vector images struggle to model complex textures. State-of-the-art methods~\cite{quint2003scalable,li2020differentiable,reddy2021im2vec,ma2022towards,chen2023editable} often rely on extensive shape coupling to approximate textures, limiting fine-grained editing and confining applications to logo design~\cite{wu2023iconshop,zhao2024vector,chen2025svgthinker} and cartoon animation~\cite{gal2024breathing,wu2024aniclipart}, rather than natural image manipulation. Another prevailing class of explicit parameterized image representations is based on Gaussian primitives~\cite{zhang2024gaussianimage,waczynska2024mirage,peng2025pixel}. However, the limited expressive capacity of individual primitives necessitates the use of a large number of stacked primitives, resulting in substantial parameter redundancy. Furthermore, since such primitives (\emph{e.g.}, Gaussian kernels) are not semantically aligned with the objects within the image, achieving fine-grained, instance-level editing remains a significant challenge.

By training low-dimensional latent embeddings, implicit parametric images~\cite{ulyanov2018deep,chen2021learning,sitzmann2020implicit,muller2022instant,vahdat2021score,rombach2022high, peebles2023scalable,zhang2024transparent,labs2025flux1kontextflowmatching} describe image in a compact way, and with large-parameter models they could even reconstruct vast numbers of examples by learning rich and realistic image priors.
However, implicit methods also exhibit notable limitations: (1) The semantic, geometric, and textural information is tightly entangled within the same latent space, making independent manipulation of specific factors highly challenging. (2) The latent space lacks a clear alignment with image semantics, lacking precise alignment between the intended editing semantics and the specific visual objects or regions, even when using carefully crafted prompts or reference conditions.
 
From the above analysis, it is evident that existing image representations struggle to balance modelling accuracy and controllability during image synthesis and manipulation. The fundamental issue lies in the entanglement of representation parameters in following dimensions: (1) Spatial entanglement: for instance, vector images rely on stacking a large number of geometric primitives to model complex textures, which hinders fine-grained editing capabilities. (2) Attribute entanglement: latent images can represent arbitrarily complex images with low-dimensional latents, yet semantic, geometric, and textural attributes are all embedded within a single parameter space, preventing targeted editing and manipulation. Any modification to a single attribute inevitably affects other attributes, and the resulting interactions are often unpredictable. As a consequence, the weak semantic alignment of current image representation methods have become a major obstacle to the further development and interactive applications of the prevailing image generation models.

\begin{figure*}
\centering
\includegraphics[width=1.0\linewidth]{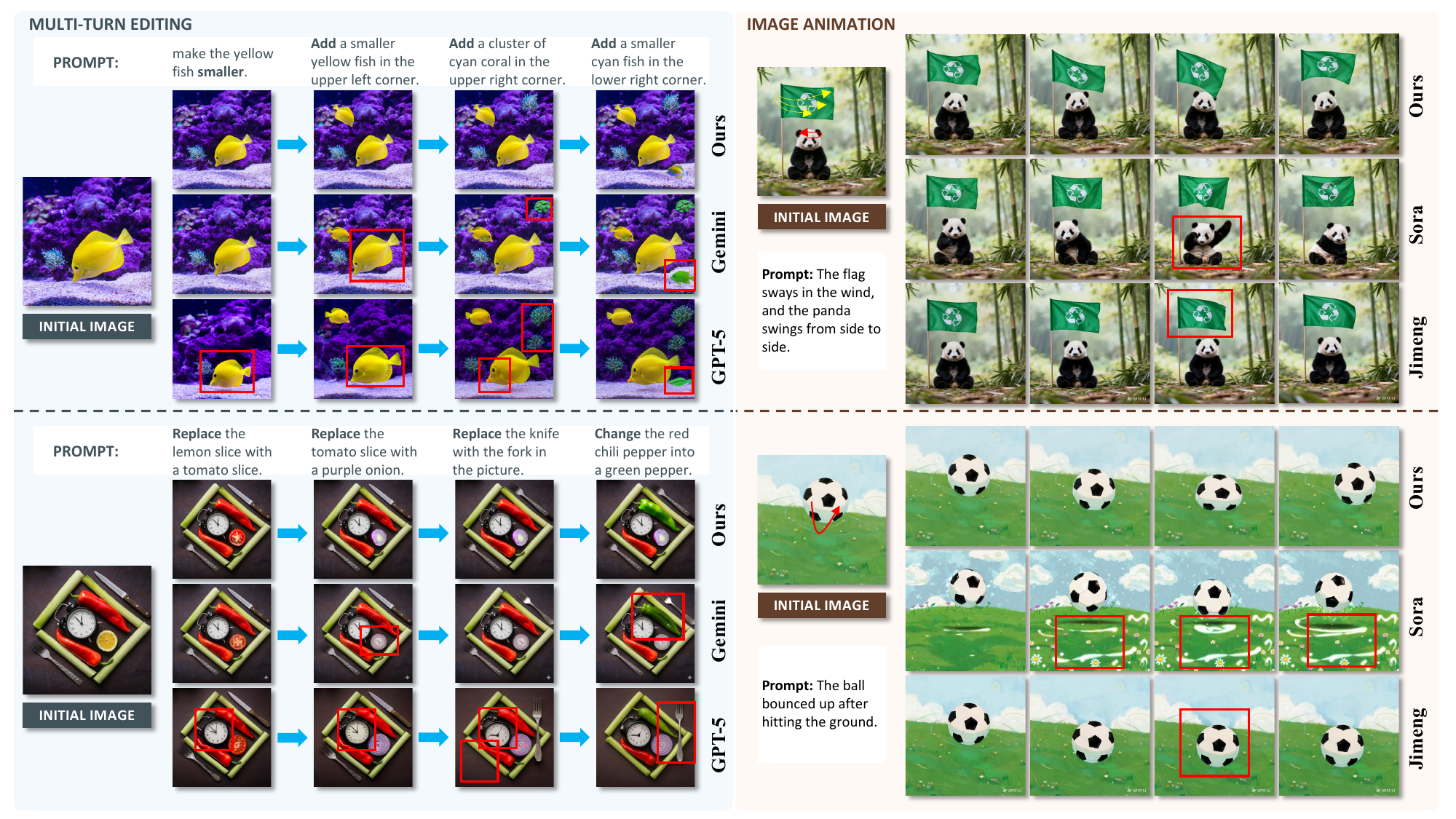}
\caption{(Left): ProxyImg enables precise instance-level multi-turn image editing, offering significantly greater control over edits compared to advanced commercial models.
(Right): ProxyImg also supports physics-based image-to-animation generation, producing animations with superior physical realism and temporal consistency while Sora and Jimeng either fail to demonstrate clear dynamic effects or exhibit significant temporal inconsistencies. ``Gemini'' means Gemini 2.5 Flash Image~\cite{comanici2025gemini} and Jimeng uses Seedance 1.0~\cite{gao2025seedance}.}
\label{fig:motivation}
\end{figure*}

To this end, we introduce a novel parametric representation that explicitly disentangles and models semantic, geometric, and textural attributes in independent, manipulable spaces. The framework operates hierarchically: it first decomposes an image into semantic layers, then constructs adaptive proxy geometries for each component using Bézier-guided geometry fitting and multi-scale internal region division/meshing, and finally embeds multi-scale implicit texture parameters/codes directly into the distributed proxy nodes, which establish a fused mapping from spatial coordinates to pixel values during image decoding, thus enabling continuous, high-fidelity reconstruction and granting unprecedented independent editability over semantic content. Our method advances the state of parametric imaging in several key aspects. Compared to explicit representations~\cite{ma2022towards,hu2024supersvg,zhang2024gaussianimage}, our semantic-layered implicit texturing significantly enhances representational capacity without relying on dense primitive stacking, thereby resolving spatial coupling and enabling component-wise manipulation. In contrast to purely implicit methods~\cite{ulyanov2018deep,sitzmann2020implicit,chen2021learning}, our geometry-grounded proxy nodes, \emph{i.e.}, encoding both envelope shapes and internal structures, provide a stable, physically-plausible support domain for the implicit mapping function. By distributing texture parameters according to geometric layout within separate per-layer spaces, our model achieves accurate texture modelling for complex natural images using only a lightweight shallow MLP for decoding.
Beyond simply storing latent codes in isolated proxy nodes, we introduce a feature plane, \emph{i.e.}, a low-resolution, spatially-aligned grid that serves as a shared, structured parameter bank. This design ensures that spatially adjacent nodes index similar features from the grid, enforcing local smoothness and enabling highly stable reconstruction with inherent spatial consistency. Moreover, by sharing the feature plane across all proxy nodes, we drastically reduce the number of parameters required for texture representation. This compact yet expressive setup not only accelerates optimization but also supports robust hole completion in occluded regions through adjacent feature propagation, all without relying on generative models.

Our representation enables low-parameter, high-fidelity reconstruction of complex natural images with scalable precision across regions. By explicitly disentangling and hierarchically aligning semantics, geometry, and texture in separate parameter spaces, the framework supports precise, component-wise editing of geometry and appearance. Extensive experiments on ImageNet, OIR-Bench, and HumanEdit show that our method outperforms existing approaches in visual quality and parameter efficiency. Furthermore, when integrated with Position-Based Dynamics, our proxy nodes allow intuitive, physically-plausible object animation via sparse control points. Thanks to distributed texture storage and a lightweight MLP renderer, the system generates high-speed animations with strong temporal consistency and visual coherence—surpassing generative-based alternatives in both controllability and efficiency.

We highlight the contributions of our work as follows:
\begin{itemize}
\item We propose a hierarchical proxy–based image parameterization that decouples image semantics, geometry, and texture, enabling high-fidelity reconstruction and efficient controllable manipulation with minimal parameters.
\item  A locality-adaptive feature assignment method further allows background layer extraction from a single image without generative models, ensuring consistency and controllability.
\item By integrating proxy representations with Position-Based Dynamics, our framework achieves real-time, physically realistic animation on low-power devices.
\item Extensive experiments show that any input image can be parameterized within three minutes with our framework, supporting diverse interactive and high-precision manipulations without extra computational cost.
\end{itemize}

\section{Related Works}
\textit{Explicit Parametric Images.} We refer to the use of explicit primitives to describe images as Explicit Parametric Image Representation. Image vectorization, which decomposes pixel-based images into a set of vectorized parameters based on graphic primitives such as B\'{e}zier shapes, has evolved substantially in recent years. Traditional approaches~\cite{xia2009patch,diebel2008bayesian} typically utilize a two-stage algorithm that involves an initial step of image segmentation, followed by targeted vectorization of the segmented regions. Diffusion curves~\cite{orzan2008diffusion,xie2014hierarchical,zhao2017inverse} and gradient meshes~\cite{sun2007image, lai2009automatic} are also frequently utilized to vectorize images. In the era of deep learning, several neural network-based methods~\cite{reddy2021im2vec,chen2023editable} are proposed inspired by the differentiable rasterization method DiffVG~\cite{li2020differentiable}.  Im2Vec~\cite{reddy2021im2vec} uses closed B\'{e}zier path as the fundamental graphic primitive. Chen~\emph{et al.}~\cite{chen2023editable} utilize several geometric primitives like triangle and rectangle to fit the image. Nevertheless, these methods all face challenges when it comes to vectorizing complex images, as predicting a large number of control points simultaneously solely based on pixel loss is very difficult. LIVE~\cite{ma2022towards} is a remarkable work that achieves layer-wise image vectorization. However, the hierarchical structure in LIVE lacks semantic information, and each layer relies on the stacking of numerous irregular shapes, making it impossible for image editing. SuperSVG~\cite{hu2024supersvg} creatively uses a ViT-based model~\cite{dosovitskiy2020image} to predict shape parameters in a feedforward manner, achieving good reconstruction accuracy for natural images. However, neural network-based approaches require significant computational resources and are difficult to generalize to arbitrary images. Inspired by 3D Gaussian Splatting (3DGS)~\cite{kerbl20233d}, recent studies~\cite{zhang2024gaussianimage,waczynska2024mirage,zhu2025large,bond2025gaussianvideo,peng2025pixel,zhang2025image,liu2025d2gv} have achieved promising results in representing 2D natural images using Gaussian primitives. Gaussianimage~\cite{zhang2024gaussianimage} is a pioneering work that represents and compresses images via 2D Gaussian Splatting, achieving high quality image reconstruction with lower memory usage and real-time rendering speed. However, such methods merely fit image textures by stacking primitives, lacking semantic and geometric alignment, and therefore remain incapable of supporting fine-grained image manipulation. This paper focuses on extracting geometric information from images using explicit primitives and enhancing the texture representation capability through distributed implicit texture encoding, thus facilitating controllable image manipulation.


\textit{Implicit Parametric Images.}
Another category of image parameterization algorithms (\emph{i.e.}, neural implicit representation~\cite{ulyanov2018deep, sitzmann2020implicit, muller2022instant,chen2021learning, chen2024image, chen2024towards}) approach the problem from an implicit representation perspective, achieving efficient coordinate-to-texture mapping through joint optimization of texture features and a decoding function. Deep image prior~\cite{ulyanov2018deep} is a pioneering work to utilize a deep network with millions of parameters to accurately fit images. SIREN~\cite{sitzmann2020implicit} uses MLPs with periodic activation functions to represent images, requiring highly redundant MLP parameters for reasonable image reconstruction. Chen~\emph{et al.}~\cite{chen2024towards} propose a shape-anchored distributed texture representation, achieving significant parameter compression. However, texture features distributed along semantic boundaries struggle to capture the internal texture structure of image components. As a result, \cite{chen2024towards} is limited to simple artistic images and performs poorly on natural images. In addition, representations mentioned above do not possess image editing capabilities. Our method introduces hierarchical geometric nodes to embed image textures in a multi-layer multi-scale manner, enabling efficient reconstruction of natural images with minimal parameters, endowed with easy image editing ability.

\textit{Image Manipulation.} In our work, we define image manipulation as a general term encompassing two tasks, \emph{i.e.}, editing and driving/animation applied to monocular images. Traditionally, explicit parametric representations, such as vectorized representations, have been widely applied to image/video editing and driving tasks~\cite{lefer2004high,dalstein2015vector,carlier2020deepsvg,mateja2023animatesvg,smith2023method,su2018live,baxter2009n,fukusato2016active,cao2023svgformer,chen2025vectorized}. AniClipart~\cite{wu2024aniclipart} is a remarkable work that integrates vectorized visual representations with text-to-video priors to generate high-quality, cartoon-style animations. However, the limited texture representational capacity of such methods confines their applicability to the manipulation of only the simplest line drawings or clipart images. In recent years, with the rapid advancement of data availability and computational power, generative models~\cite{rombach2022high,zhang2023adding,labs2025flux1kontextflowmatching} based on latent image representations achieve remarkable progress in natural image editing~\cite{brooks2023instructpix2pix,sheynin2024emu,yang2023imagebrush,huang2024smartedit,xie2023smartbrush,ge2024seed,zhao2024ultraedit,yu2025anyedit} and driving~\cite{wu2023tune,guo2023animatediff,yang2024cogvideox,kong2024hunyuanvideo,wan2025wan,blattmann2023stable,dai2023animateanything}. Latent Diffusion Models (LDMs)~\cite{rombach2022high} apply diffusion processes in the latent space of pretrained autoencoders and achieves high-quality and efficient image synthesis with strong conditioning flexibility and greatly reduced computational cost, which provides a fundamental basis for instruction-guided image editing tasks. AnimateDiff~\cite{guo2023animatediff} introduces a plug-and-play motion module that enables personalized text-to-image models to generate temporally smooth animations without model-specific tuning, achieving high visual fidelity and motion diversity. However, such methods are constrained by the misalignment between the latent and pixel spaces, making it difficult to achieve fully aligned and controllable image editing and motion driving according to user intent. This paper aims to propose a method that disentangles and parameterizes multiple attributes of input images, enabling controllable image editing and animation.

\begin{figure*}
\centering
\includegraphics[width=1.0\linewidth]{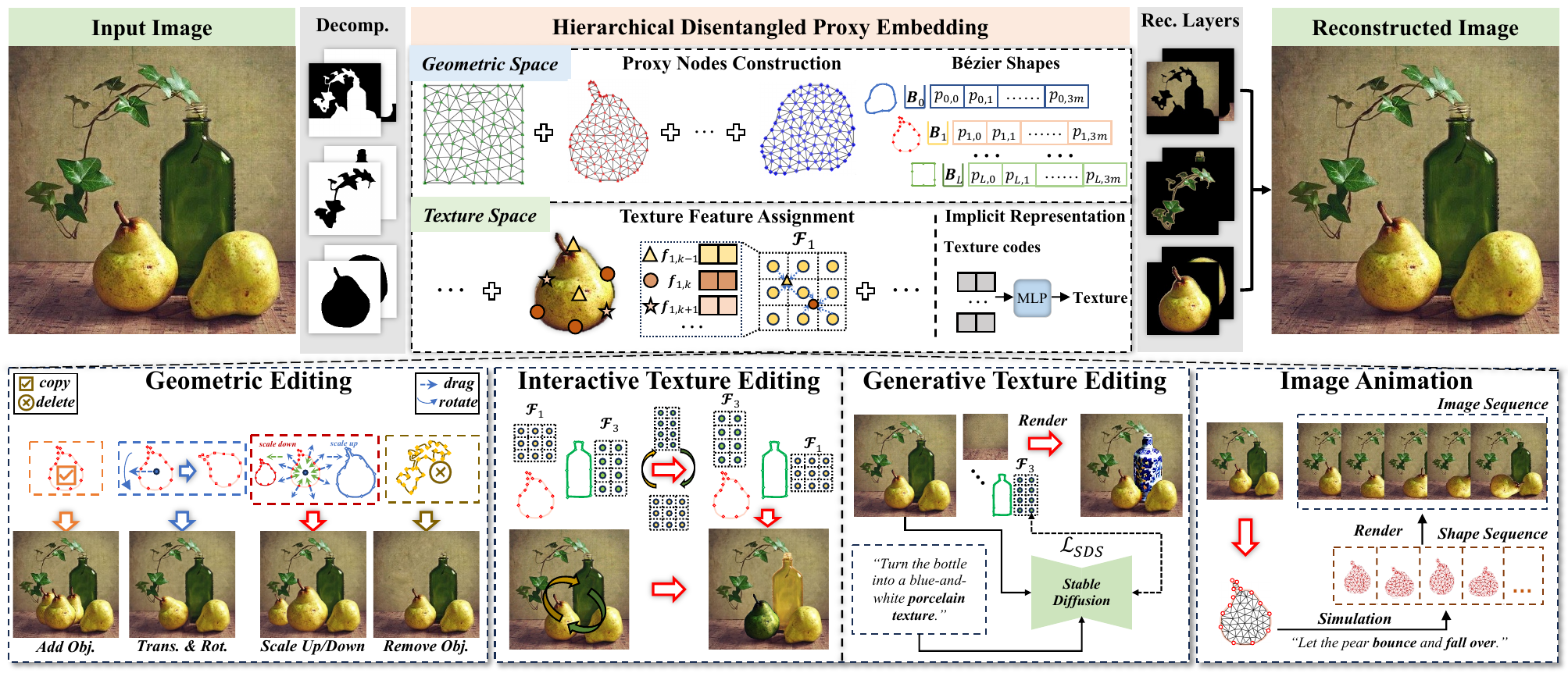}
\caption{\textbf{Overview of our framework.} We propose a novel vectorized image representation to embed multi-scale image texture features on multi-layer hierarchically related geometric control points. With such representation, the texture at arbitrary position in the continuous image domain can be decoded with a geometry-aware interpolation method and a lightweight decoding function. Building upon our representation, we can enable a variety of controllable and precise image manipulations, such as geometric editing, texture editing, and image animation, with only minimal computational cost.}
\label{fig:framework}
\end{figure*}

\section{Methodology}
\textbf{Method Overview}. We introduce a novel parametric representation that decouples structure and texture by constructing a hierarchical proxy embedding for each semantically segmented instance. As illustrated in Fig.~\ref{fig:framework}, our key innovation lies in a dual-decoupling design, which disentangles texture from geometry within each semantic layer. First, we model instance geometry using both adaptive Bézier curves that capture precise boundaries and a multi-scale set of internal proxy nodes that explicitly represent its internal structural hierarchy. Second, we embed optimizable texture features onto these distributed geometric proxies, which, combined with a lightweight decoder, enable a frequency-aware implicit texture representation that adapts to the spatially varying spectral characteristics of natural images. This explicit separation of shape (structure) and texture (appearance) within a unified proxy-based framework forms the foundation for highly controllable and efficient image editing.

\textbf{Mathematical Formulation}. Accordingly, an input image $\boldsymbol{I} \in \mathbb{R}^{W \times H \times 3}$ is parameterized into a set of \emph{layer-wise proxy-based representations} and a shared lightweight decoder. Formally, the image is represented as  
\begin{equation}
\boldsymbol{I} \sim \big\{ \{\boldsymbol{R}_{1}, \boldsymbol{R}_{2}, \dots, \boldsymbol{R}_{L}\}, \, \theta \big\},
\label{eq:def1}
\end{equation}
where $L$ denotes the number of semantic layers, and each layer \(\boldsymbol{R}_i\) is defined by  
$\boldsymbol{R}_i = \{ \boldsymbol{B}_i, \; \boldsymbol{G}_i, \; \boldsymbol{F}_i \}$. 
Here, $\boldsymbol{B}_i$ denotes the Bézier shape capturing the layer boundary, $\boldsymbol{G}_i$ represents the hierarchical set of internal proxy nodes that model the layer’s geometry, and $\boldsymbol{F}_i$ corresponds to the texture features embedded on these geometric proxies. The trainable function $\phi_\theta$ acts as a coordinate-to-RGB decoder that synthesizes pixel colors from the interpolated features, enabling a continuous and editable image representation.

\textbf{Applications and Edits}. Our representation directly enables three core editing operations through simple mathematical manipulations and re-rendering of its components. 

\textbf{Geometric editing }is achieved by applying affine transformations directly to the proxy nodes along layer boundary as:
\begin{equation}
\label{eq:shapeedit}
\hat{\boldsymbol{B}_{i}} = \boldsymbol{T} (\boldsymbol{B}_{i} - \boldsymbol{c}_i) + (\boldsymbol{c}_i + [t_x,t_y]^{T}),
\end{equation}
where $\boldsymbol{c}_i$ denotes the geometric center of the layer. $[t_x,t_y]$ denote the overall translation parameter. $\boldsymbol{T} \in \mathbb{R}^{2 \times 2}$ is the affine transformation matrix, which allows for rotation, and scaling of instances. In addition, we can perform instance addition or removal by globally copying or deleting the proxy nodes of a specific layer. It is worth noting that users only need to manipulate the Bézier control points, and the positions of the internal proxy nodes will be automatically reconstructed based on the positions of the Bézier control points. At the same time, the distributed texture storage method ensures the stability of the texture throughout the process.

Texture editing is performed by modifying the feature embeddings on relevant nodes. Our approach supports two texture editing modes: (a) \textbf{Interactive Texture Editing:} Users can directly assign the texture of instance $j$ to instance $i$ in the image by simply copying the texture features, which can be denoted as:
\begin{equation}
\label{eq:textureedit1}
\hat{\boldsymbol{R}}_{i} = \{\boldsymbol{B}_i, \boldsymbol{G}_i, \hat{\boldsymbol{F}}_i\Leftarrow\boldsymbol{F}_j\},
\end{equation}
where $\Leftarrow$ means using the proxy nodes of instance $i$ to interpolate features from the texture space of instance $j$.
(b) \textbf{Generative Texture Editing:} We leverage a pretrained image generation model to optimize the texture features of the target semantic layer via Score Distillation Sampling (SDS)~\cite{poole2022dreamfusion}, enabling the synthesis of novel textures not present in the observed image (for example, replacing the fur of a dog with the texture of a tiger's skin):
\begin{equation}
\nabla_{\Theta}\mathcal{L}_{\text{SDS}}=\mathbb{E}_{t,\boldsymbol{\epsilon}}\left[w(t)\left(\boldsymbol{\hat{\epsilon}}\left(\boldsymbol{z}_{t} ;t,\boldsymbol{y}\right)-\boldsymbol{\epsilon}\right)\frac{\partial\boldsymbol{z_{t}}}{\partial\Theta}\right],
\label{eq:sds}
\end{equation}
where $\Theta=\{\theta,\boldsymbol{F}_{i}\}$ denotes the re-optimized parameters, $w(t)$ is a weighting function, $\boldsymbol{y}$ is the editing prompt, $\boldsymbol{z}_{t}$ is the latent code of the rendered image, with a noise level $t$. Optimization process is repeated, with texture features converging toward rendering results that match the editing prompt.

Finally, image animation is realized by driving the proxy nodes with Position-Based Dynamics (PBD) as
\begin{equation}
\label{eq:animate}
\{\hat{\boldsymbol{B}}_i, \hat{\boldsymbol{G}}_i\} = Sim(\{\boldsymbol{B}_i, \boldsymbol{G}_i\}),
\end{equation}
where $Sim$ refers to the physical simulation of proxy nodes based on user interaction, which simulates physically plausible motion, while the attached features ensure temporal consistency in rendering.

\subsection{Instance-level Geometry Representation: Hierarchical Spatial Proxy Nodes Construction}
\label{sec:geo}
\subsubsection{Semantic Instance Layer Decomposition}
To enable independent and interference-free editing of image components, our first step is to decompose the image into distinct semantic instances. We first employ the Segment Anything Model (SAM)~\cite{kirillov2023segment} to generate a series of separate image regions: $\small\{\boldsymbol{M}_{1}, \boldsymbol{M}_{2}..., \boldsymbol{M}_{L}\} = SAM(\boldsymbol{I})$,
where $\small\boldsymbol{M}_{i}\in \{0,1\}^{W\times H}$ represents the $i-$th image layer (as mask). Next, we utilize the Depth Anything Model~\cite{yang2024depth} to generate the depth map of the input image and calculate the average depth value $\bar d_i$ for each masked region $\small\boldsymbol{M}_{i}$. We then sort all layers according to their average depth, assigning the layer with the largest depth value as the background layer, while treating the remaining layers as foreground layers, yielding $L-1$ foreground layers and one background layer as
\begin{equation}
    \boldsymbol{I} \rightarrow \{\boldsymbol{M}_1^{f},...,\boldsymbol{M}_{L-1}^{f},\boldsymbol{M}^{b}\}.
    \label{eqn:decompose}
\end{equation}
The above decomposition enables independent manipulation/rendering of the background and various foreground semantic components. We directly set the background layer as a complete mask with the same size as the source image (\emph{i.e.}, $\boldsymbol{M}^{b}=\mathbf{1}^{W\times H}$), in order to prevent holes from appearing during image manipulation tasks, \emph{i.e.}, object relocation, deformation, or animation \emph{etc}.

\subsubsection{Adaptive Instance Shape Boundary Proxy Nodes Generation}
For each image layer $\small\boldsymbol{M}_{i}$, the goal is to efficiently approximate its boundaries using a single B\'{e}zier shape (\emph{i.e.}, Image Vectorization~\cite{ma2022towards,hu2024supersvg,chen2024towards,chen2025easy,chen2025vectorized}). This enables the editing and manipulation of an entire semantic layer through the control of a few key points. 
Exiting methods~\cite{reddy2021im2vec,ma2022towards,chen2023editable,chen2024towards} are restricted by a predetermined count of shape boundary segments, imposing a trade-off between descriptive capacity and computational efficiency.
In contrast, we model each layer as a single shape with an adaptive control point optimization strategy, thus we can fit any shape with an arbitrary number of control points adaptively and avoid shape stacking, which not only enhances the accuracy and parameter efficiency of B\'{e}zier shapes, but also facilitates easy editing of image semantic components.

Specifically, we first detect the shape boundary points $\small\mathbb{E}_{i}=\{\boldsymbol{e}_{k}\}$ and sample $3m$ points from them as initialization of control points for an $m-$segment B\'{e}zier shape $\small\boldsymbol{B}_{i}=\{\boldsymbol{s}_{i1},...,\boldsymbol{s}_{im}\}$, where each segment $\boldsymbol{s}_{ij}$ is parameterized by four control points $\small\{\boldsymbol{p}_{ij}^{0}, \boldsymbol{p}_{ij}^{1}, \boldsymbol{p}_{ij}^{2}, \boldsymbol{p}_{ij}^{3}\}$ and we can sample points in this segment uniformly by:
\begin{equation}
    \mathbb{B}_{ij} = \{\boldsymbol{b}_{ij}(t)=\sum_{o=0}^{3} \binom{3}{o} (1 - t)^{3 - o} t^o \boldsymbol{p}_{ij}^{o}\},
\end{equation}
where $t\in [0,1]$ denotes the sampling coefficient. The points sampled across the entire shape can be denoted as $\small\mathbb{B}_{i}=\cup_{j=1}^m \mathbb{B}_{ij}$.
We then calculate the Chamfer distance between $\mathbb{E}_{i}$ and $\mathbb{B}_{i}$ to optimize control point positions:
\begin{equation}
\mathcal{L}^{cd}_{i}=\sum_{\boldsymbol{b}\in \mathbb{B}_{i}}\min_{\boldsymbol{e}\in \mathbb{E}_{i}} \|\boldsymbol{b}-\boldsymbol{e}\|^2_2+\sum_{\boldsymbol{e}\in \mathbb{E}_{i}}
\min_{\boldsymbol{b}\in \mathbb{B}_{i}} \|\boldsymbol{e}-\boldsymbol{b}\|^2_2. 
\label{eq:loss_CD}
\vspace{-0.1cm}
\end{equation}
After optimizing for a number of epochs, we compute a per-segment error for the B\'{e}zier shape:
\begin{equation}
    err(\boldsymbol{s}_{ij}) = \sum_{t} ||\boldsymbol{b}_{ij}(t) - \boldsymbol{e}_{ij}(t)||_2^2,
    \vspace{-0.2cm}
\end{equation}
where $\small\boldsymbol{e}_{ij}\in \mathbb{E}_{i}$ represents the boundary point that is closest to the shape point $\small\boldsymbol{b}_{ij}$. For each segment $\boldsymbol{s}_{ij}$ with $\small err(\boldsymbol{s}_{ij}) > \tau$, we sample three points $\small\{\boldsymbol{p}_{ij}^{11}, \boldsymbol{p}_{ij}^{12},\boldsymbol{p}_{ij}^{13}\}$ along the segment and insert them into the control point sequence, thus splitting the segment into two segments $\boldsymbol{s}_{ij1}$ and $\boldsymbol{s}_{ij2}$, which are parameterized by control points $\small\{\boldsymbol{p}_{ij}^{0}, \boldsymbol{p}_{ij}^{1}, \boldsymbol{p}_{ij}^{11},\boldsymbol{p}_{ij}^{12}\}$ and $\small\{\boldsymbol{p}_{ij}^{12}, \boldsymbol{p}_{ij}^{13}, \boldsymbol{p}_{ij}^{2},\boldsymbol{p}_{ij}^{3}\}$, respectively. We repeat the above process until the error of any segment is less than $\tau$ and then obtain the optimized $\small\boldsymbol{B}_{i}$.

\subsubsection{Multi-scale Internal Geometrical Proxy Nodes Generation}
Merely fitting external boundaries is insufficient to represent the complex geometric internal structures of an instance, especially the object is undergoing non-planar motion with significant internal shape deformation/distortion~\cite{ma2022towards,chen2024towards}. We therefore introduce internal proxy geometric nodes based on an interrelated multi-scale triangulation method within each semantic layer (instance region) to establish a complete structural support, efficiently with fewer parameters.

Specifically, we first employ the Triwild algorithm~\cite{hu2019triwild}, using the B\'{e}zier curves as boundary constraints, to recursively triangulate the interior of the layer. Note that the relative edge length $l_r$ is a key parameter that controls the density of the generated triangles during the triangulation process. Considering that the texture of background layers tends to be smoother, while foreground layers often exhibit richer textures, we assign different base edge lengths for background and foreground layers respectively, \emph{i.e.} $l_r^{b}$ and $l_r^f$. 

The first layer proxy geometry (denoted as \textit{hierarchy-1}) is generated as follows:
\begin{equation}
    \boldsymbol{G}_{i}^{1} =\{\boldsymbol{V}_{i}^{1}, \boldsymbol{T}_{i}^{1}\} = Triwild(\boldsymbol{M}_{i}, \boldsymbol{B}_{i}, l_r^{*}),
    \vspace{-0.1cm}
\end{equation}
where $\small\boldsymbol{V}_{i}^{1}$ denotes the collection of the \textit{hierarchy-1} proxy vertices and $\small\boldsymbol{T}_{i}^{1}$ represents the proxy triangle elements. 
To achieve a multi-scale representation, we then dynamically refine these level-1 triangles: regions with higher geometric information density should be represented with finer triangles. We use gradient magnitude—extracted via a Sobel filter—as the measure of information density. Specifically, we first compute a gradient map for the input image $\small\boldsymbol{M}{i}$. For each triangle $\small\boldsymbol{t}_{ij}^{1} \in \boldsymbol{T}_{i}^{1}$, we calculate the average gradient magnitude over the pixels it covers. If the triangle spans more than one pixel and its average gradient exceeds a predefined threshold (typically set to the mean gradient magnitude of the entire image), we refine it by sampling a new vertex on each of its three edges, subdividing it into four smaller triangles. The above operation is performed in parallel for all triangles yields the \textit{hierarchy-2} proxy geometric nodes, which can be described as:
\begin{equation}
    \boldsymbol{G}_{i}^{2} = \{\boldsymbol{V}_{i}^{2}, \boldsymbol{T}_{i}^{2}\} = Refine(\boldsymbol{V}_{i}^{1}, \boldsymbol{T}_{i}^{1}).
\end{equation}
We perform the above process recursively to obtain proxy geometric nodes with multiple hierarchical levels for each image layer $\small\boldsymbol{M}_{i}$. The generated hierarchical proxy geometry can be denoted as:
\begin{equation}
    \boldsymbol{G}_{i}^{*} = \cup_{j=1}^{h^{*}}\boldsymbol{G}_{i}^{*,j},
\end{equation}
where $\small *\in \{b,f\}$ denotes foreground/background layers. $\small\boldsymbol{G}_{i}^{*,j}=\{\boldsymbol{V}_{i}^{*,j}, \boldsymbol{T}_{i}^{*,j}\}$ denotes the \textit{hierarchy-j} proxy geometry for $\boldsymbol{M}_i$. In our experiments, we set $\small h^{b}=2$ and $\small h^{f}=3$.

\subsection{Distributed Proxy Representation of Image Texture}
\label{sec:texture}
\begin{figure*}
\centering
\includegraphics[width=1.0\linewidth]{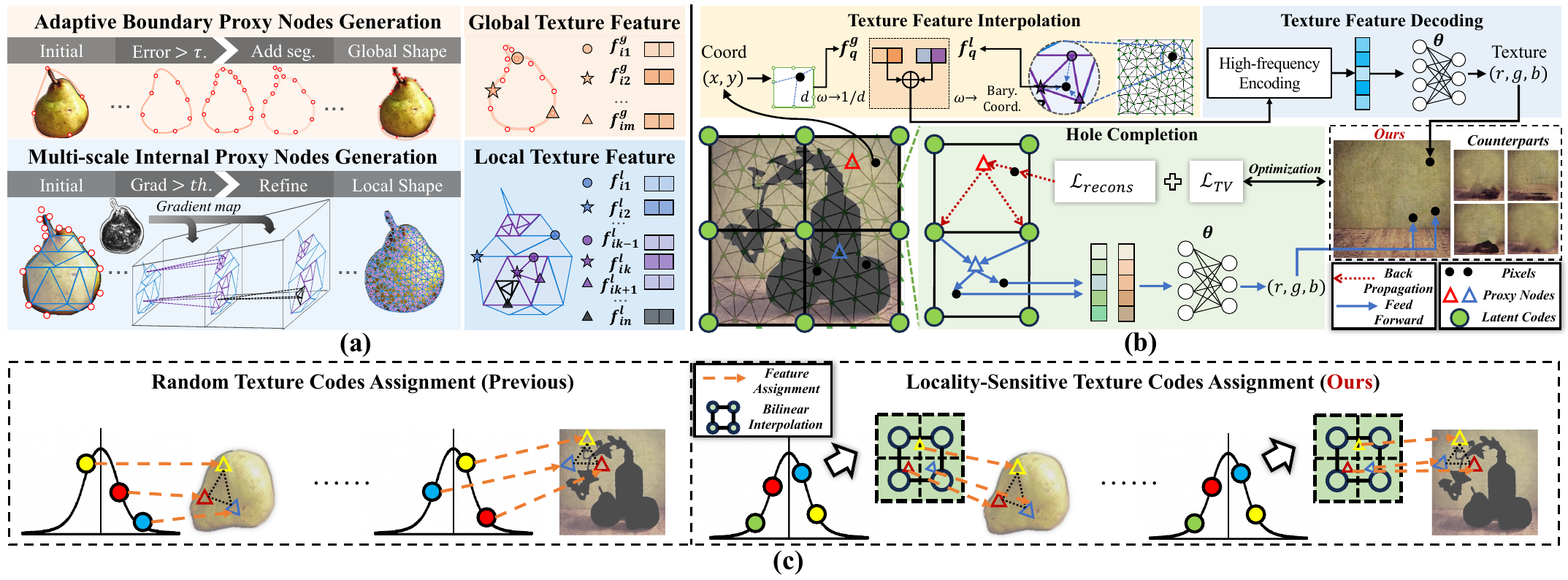}
\caption{(a). Diagram of proxy nodes construction and distributed embedding of image textures. (b). Schematic of the coordinate-to-texture network training process based on proxy embeddings and illustration of background hole completion using a learnable spatial feature grid (represented with green dots). (c). Illustration of the differences between the proposed feature assignment method and the commonly used random feature assignment. Compared with prior methods, the proposed algorithm effectively ensures distributional consistency of local texture codes, which not only enhances texture stability during image manipulation but also accelerates the parameter optimization process. Please zoom in and refer to the legends for more details.}
\label{fig:framework2}
\end{figure*}

Following our prior work on distributed implicit representations~\cite{chen2024towards, chen2025easy,chen2025vectorized}, we attach texture features to the geometric proxy nodes (control points) and decode them via a lightweight coordinate-to-RGB network, achieving efficient and compact texture encoding.

Specifically, to capture both the global texture information of layer $\small \boldsymbol{M}_{i}$ as well as the rich local textures within the layer, we sample 1) global texture embedding points from the boundary B\'{e}zier shape $\small \boldsymbol{B}_{i}$ as texture control points, \emph{i.e.}, uniformly sampling one point in each segment yielding feature codes $\small [\boldsymbol{f}^{g}_{i1},...,\boldsymbol{f}^{g}_{im}]$; and 2) internal/local texture embedding points at all the multi-scale proxy vertices (\emph{i.e.}, $\small \boldsymbol{V}_{i}$) within the layer: $\small [\boldsymbol{f}^{l}_{\boldsymbol{v}_{i}^{1}},...,\boldsymbol{f}^{l}_{\boldsymbol{v}_{i}^{n}}]$,
where $\small\boldsymbol{v}_{i}^{*}$ means the vertices of all the hierarchical triangles mentioned in Sec.~\ref{sec:geo} and $n$ denotes the number of vertices.

Consequently, we parameterize the texture information of each image as hierarchical texture feature vectors (\emph{i.e.}, $\small\boldsymbol{F}_{i}=[\boldsymbol{F}^{g}_{i},\boldsymbol{F}^{l}_{i}]$)
embedded at geometric control points, along with a lightweight decoding function $\phi_\theta:\boldsymbol{f_x}\mapsto\boldsymbol{c}$ to map features to texture values, where $\boldsymbol{x}$ is the queried coordinate and $\small\boldsymbol{f_x}$ denotes the corresponding feature vector at the queried point, which can be obtained through a geometry-aware interpolation method. Specifically, for arbitrarily positioned query point in $\small\boldsymbol{M}_{i}$ with coordinate $\boldsymbol{x}$, we follow~\cite{chen2024towards} to use inverse distance weighting method to interpolate its global texture feature from features stored at B\'{e}zier control points:
\begin{equation}
\label{eq:global_feat}
    \boldsymbol{f}_{\boldsymbol{x}}^g = \sum_{j=1}^{m} \frac{w(\boldsymbol{x}, \boldsymbol{p}_{ij}^{s})}{\sum_{j=1}^{m} w(\boldsymbol{x}, \boldsymbol{p}_{ij}^{s})} \boldsymbol{f}_{ij}^{g},
\end{equation}
where $\small\boldsymbol{p}_{ij}^{s}$ means the texture control point sampled from the $j$-th segment. $\small w(\cdot, \cdot)$ is the inverse distance weighting function. For local texture features, we first use the barycentric coordinate algorithm to locate the triangle element $\small \boldsymbol{t}_{ij}=[\boldsymbol{v}_{ij}^{1},\boldsymbol{v}_{ij}^{2},\boldsymbol{v}_{ij}^{3}]$ that covers point $\boldsymbol{x}$ within all the hierarchical proxy triangles (\emph{i.e.}, $\small \boldsymbol{T}_{i}$).
We then use barycentric coordinate interpolation to obtain the local texture feature of point $\boldsymbol{x}$:
\begin{equation}
\label{eq:local_feat}
    \boldsymbol{f}^{l}_{\boldsymbol{x}} = \sum_{k=1}^{3} \lambda_{\boldsymbol{x}}^{ij,k} \boldsymbol{f}^{l}_{\boldsymbol{v}_{ij}^{k}},
\end{equation}
where $\small (\lambda_{\boldsymbol{x}}^{ij,1}, \lambda_{\boldsymbol{x}}^{ij,2}, \lambda_{\boldsymbol{x}}^{ij,3})$ denotes the barycentric coordinate vector of point $\boldsymbol{x}$ with respect to triangle element $\small\boldsymbol{t}_{ij}$ and $\small\boldsymbol{f}^{l}_{*}$ means the local feature vectors at each vertex of $\small\boldsymbol{t}_{ij}$. Thus, for any query point $\boldsymbol{x}$, we can interpolate its texture features: $\small\boldsymbol{f_x} = [\boldsymbol{f}^{g}_{\boldsymbol{x}},\boldsymbol{f}^{l}_{\boldsymbol{x}}]$.

\subsubsection{Locality-sensitive Texture Encoding}
Assigning independent, randomly initialized texture features to each proxy node overlooks spatial continuity and leads to two fundamental issues: it makes optimization unstable, hinders parameter efficiency as storage scales with node count, and causes texture artifacts during editing due to abrupt feature changes under geometric deformation.

To enforce local smoothness and achieve compactness, we depart from per-node feature storage. Instead, we introduce a shared, low-resolution feature grid that acts as a spatial lookup table. The texture code for any node is obtained by retrieving and interpolating within this grid based on its normalized coordinates. This design inherently ensures that nearby nodes retrieve similar features, providing a stable local support for texture synthesis and editing while drastically reducing the number of parameters to the size of the grid.

\textbf{Locality-Sensitive Texture Codes Assignment}. For each semantic layer $\small \boldsymbol{M}_{i}$, we design a \emph{learnable spatial feature grid} $\boldsymbol{\mathcal{F}}_i \in \mathbb{R}^{W_i \times H_i \times D}$ (note that the feature grid for different layer should be different and independent in order to capture multi-scale information), which serves as a shared, low-resolution lookup table of texture prototypes. The grid spatially tiles the layer's bounding box, where each cell stores a $D$-dimensional feature vector capturing the local texture statistics of that region. Then texture vectors of layer $i$ (\emph{i.e.}, $\boldsymbol{F}_i$) can be sampled from $\boldsymbol{\mathcal{F}}_i$. Thus the texture feature for a proxy node at continuous coordinate $\boldsymbol{v}_{ij}$ is not stored directly but is dynamically retrieved via a coordinate mapping followed by bilinear interpolation. Specifically, we first normalize $\boldsymbol{v}_{ij}$ within the minimum bounding box of its corresponding layer, and then project it into the feature map space, which can be denoted as:
\begin{align}
\label{eq:assignment1}
    \boldsymbol{v}_{ij}^{norm} &= [\frac{{v}_{ij,x} - {m}_{i,x}^{00}}{{m}_{i,x}^{11} - {m}_{i,x}^{00}},\frac{{v}_{ij,y} - {m}_{i,y}^{00}}{{m}_{i,y}^{11} - {m}_{i,y}^{00}}], \\
\label{eq:assignment2}
     \boldsymbol{f}_{\boldsymbol{v}_{ij}} &= \sum_{k}\alpha(\boldsymbol{v}_{ij}^{norm},\boldsymbol{q}_{\boldsymbol{v}_{ij}}^{k})\boldsymbol{f}_{\boldsymbol{q}_{\boldsymbol{v}_{ij}}^{k}}, 
\end{align}
where $\small \boldsymbol{m}_{i}^{00}=[m_{i,x}^{00}, m_{i,y}^{00}]$ and $\small \boldsymbol{m}_{i}^{11}=[m_{i,x}^{11}, m_{i,y}^{11}]$ define the bounding box of layer $i$, $\small \boldsymbol{q}_{\boldsymbol{v}_{ij}}^{k}$ and $\small \boldsymbol{f}_{\boldsymbol{q}_{\boldsymbol{v}{ij}}^{k}}$ denote the normalized coordinates and feature vectors of nearest latent codes of $\small \boldsymbol{v}_{ij}$ in the four quadrant sub-spaces of the feature map $\small \boldsymbol{\mathcal{F}}_{i}$, with $\small k\in \{00,01,10,11\}$ indexing the top-left, top-right, bottom-left, and bottom-right regions, respectively. $\alpha(\cdot,\cdot)$ represent the bilinear interpolation weights. It is worth noting that Eqn.~\ref{eq:assignment1} and Eqn.~\ref{eq:assignment2} apply to the feature assignment of all proxy nodes, including both edge (global) nodes and internal (local) nodes. This continuous interpolation guarantees that \emph{spatially adjacent nodes receive similar features}, creating a locally smooth embedding field ideal for stable optimization and coherent editing. Furthermore, the parameter count is fixed to the grid size $|W_i \times H_i \times D|$ and independent of the number of proxy nodes, achieving both efficiency and representation compactness.

\textbf{Map Size Computation}. The size of the feature map, which determines the spatial extent represented by each latent code, is a critical design parameter. For foreground layers with complex and variable textures, we employ a larger feature map to increase the number of latent codes, allowing each to model finer local patterns. Specifically, the feature map size ($W_i^f, \text{H}_i^f$) for a foreground layer is adapted to its bounding box as follows:
\begin{equation}
\label{eqn:hole1}
    [W_i^f, H_i^f] = (\boldsymbol{m}_{i}^{11} - \boldsymbol{m}_{i}^{00}) / o_i,
\end{equation}
where $s_i$ is used to balance the representational capacity and the number of parameters and we set $o_i=8$ in our experiments.
For the background layer, a key challenge is to complete occluded regions (holes) where proxy nodes lack direct pixel supervision. To propagate texture information from visible areas into these holes, we require each latent code in the feature map to cover a sufficiently broad spatial region. Therefore, we \textbf{deliberately keep the background feature map small}—just large enough to maintain basic representational capacity while ensuring strong spatial coupling between nodes inside and outside holes. Denoting the proxy nodes inside the largest hole as $\mathbb{N}$ and those in visible background areas as $\mathbb{P}$, we compute the maximum spatial gap as:
\begin{align}
    \delta_x &= \max_{\boldsymbol{v}_n \in \mathbb{N}} \min_{\boldsymbol{v}_p \in \mathbb{P}} ||\boldsymbol{v}_{n,x} - \boldsymbol{v}_{p,x}||, \\
    \delta_y &= \max_{\boldsymbol{v}_n \in \mathbb{N}} \min_{\boldsymbol{v}_p \in \mathbb{P}} ||\boldsymbol{v}_{n,y} - \boldsymbol{v}_{p,y}||.
\end{align}
Thus size of feature map for background layer should be defined as:
\begin{equation}
\label{eqn:hole2}
    W_i^{b} < W/\delta_x, ~~H_i^{b} < H/\delta_y,
\end{equation}
where $W$ and $H$ denote the size of the input image. Therefore, to balance hole completion and image representation capability, we set $W_i^{b}$ and $H_i^{b}$ as $max[8,\lfloor W/\delta_x\rfloor]$ and $max[8,\lfloor H/\delta_y\rfloor]$, respectively.


\subsection{Parameter Optimization}
For geometric parameters, the coordinates of B\'{e}zier control points and multi-scale triangles are obtained as stated in Sec.~\ref{sec:geo}. For texture parameters (\emph{i.e.}, feature maps $\boldsymbol{F}=\{\boldsymbol{F_i}|i=1,...,L\}$ and decoding function $\phi_\theta$), we first assign texture features to all the proxy nodes according to Eqn.~\ref{eq:assignment1} and Eqn.~\ref{eq:assignment2}. Then we follow~\cite{chen2021learning, chen2024towards} to treat each pixel $\small[i,j]$ as a coordinate $\boldsymbol{x}_{ij}$ in normalized continuous image space. For each point, we search its corresponding layer according to the layer order defined in Eqn.~\ref{eqn:decompose} and then obtain its hierarchical features $\small\boldsymbol{f}_{\boldsymbol{x}_{ij}}$ as described in Eqn.~\ref{eq:global_feat} and Eqn.~\ref{eq:local_feat}. It is noted that we only need to assess the positional relationship between the pixel coordinates and the Bézier shapes, eliminating the need to store a binary mask for each layer. Then we can predict the texture value at pixel $\small[i,j]$ with the decoding function $\phi_{\theta}$. To more effectively capture high-frequency details in images while decoupling image texture from pixel coordinates for stable and controllable image manipulation, we abandon the conventional position encoding technique and instead apply high-frequency encoding to the pixel features, inspired by prior work~\cite{dou2023multiplicative}:
\begin{equation}
    \widetilde{\boldsymbol{f}}_{\boldsymbol{x}_{ij}} = \mathcal{U}(\boldsymbol{f}_{\boldsymbol{x}_{ij}}),
\end{equation}
where $\small\mathcal{U}$ denotes the parameter-free encoding function as defined in~\cite{mildenhall2021nerf}. Hence the output image $\small\hat{\boldsymbol{I}}$ can be described as:
\begin{equation}
    \hat{\boldsymbol{I}}[i,j] = \phi_\theta(\widetilde{\boldsymbol{f}}_{\boldsymbol{x}_{ij}}).
\end{equation}
Then we can use pixel-wise error to optimize the texture features of all image layers (\emph{i.e.}, $\small\boldsymbol{F}$) and the parameters $\theta$ of the decoding function simultaneously:
\begin{equation}
    \mathcal{L}_{recons} = ||\boldsymbol{I} - \boldsymbol{\hat{I}}||_{2}^{2}.
\end{equation}

Additionally, considering that the completion of occluded regions (holes) in the background is crucial for subsequent image manipulation tasks, we apply extra rendering and optimization to the feature map of the background layer. The objective of optimizing the background layer is to ensure that the completed background layer is precise and seamless, especially around the edges of the holes, which should blend smoothly and naturally with the original background. To achieve this, we introduce a Total Variation (TV) Loss~\cite{mahendran2015understanding} to regularize the background layer. More concretely, we obtain the entire background layer $\boldsymbol{I}_B$ by rendering the proxy nodes of the background and their corresponding feature map in each optimization step. We then apply dilation and erosion to the mask of each foreground layer, thereby obtaining the boundary regions corresponding to each hole on the background:
\begin{equation}
    \boldsymbol{M}_{i}^{f,edge} = \textit{dil}(\boldsymbol{M}_{i}^{f}) \oplus \textit{ero}(\boldsymbol{M}_{i}^{f}),
\end{equation}
where $\textit{dil}(\cdot)$ and $\textit{ero}(\cdot)$ denote mask dilation and erosion, respectively. $\oplus$ means the XOR operation between two binary masks. The TV Loss is then applied to the entire background layer as well as the boundary regions of all holes, which can be denoted as:
\begin{equation}
    \mathcal{L}_{tv} = TV(\boldsymbol{I}_B) + \beta \sum_{i=1}^{L-1}  TV(\boldsymbol{I}_B \cdot \boldsymbol{M}_{i}^{f,edge}),
\end{equation}
where $\beta$ is set to 10 to enhance the handling of hole boundaries and make the transition in the occluded background regions more natural. $TV(\cdot)$ is the Total Variation Loss as defined in~\cite{mahendran2015understanding}:
\begin{equation}
    TV(\mathbf{x}) = \sum_{i,j}((x_{i,j+1} - x_{ij})^2 + (x_{i+1,j} - x_{ij})^2)^{\frac{1}{2}}. 
\end{equation}
The overall optimization objective can be denoted as:
\begin{equation}
    \min_{\{\boldsymbol{F},\theta\}} \mathcal{L}_{recons} + \gamma\mathcal{L}_{tv},
\end{equation}
where $\gamma$ is used to balance the magnitudes of the losses.

\section{Experiments}
In this section, we evaluate the proposed representation using two main tasks: image representation and image manipulation. First, to assess the image representation capability of the proposed method, we compare its image reconstruction accuracy and parameter count with current state-of-the-art image representation algorithms, including both explicit and implicit representations. Notably, to demonstrate the potential of our representation in image parameter compression, we apply a simple quantization algorithm to our representation parameters and compare them with dedicated image compression algorithms. For image manipulation, we define three major tasks: object geometry editing, object texture editing, and image animation. Each of these tasks is compared with advanced methods specialized in the respective task. It is important to note that once any input image is parameterized through the proposed representation framework, it can be directly used for all tasks without the need for introducing additional parameters.

\subsection{Experimental Setups}
\noindent\textbf{Implementation Details.} We initialize $4-$segment B\'{e}zier shape for each layer, \emph{i.e.}, $\small m=4$. During adaptive B\'{e}zier shape fitting, we optimize for a total of 2000 epochs and calculate the per-segment error every 500 epochs. We set the error threshold $\small \tau=2e-3$. For Triwild algorithm, we set base edge lengths as $\small l_r^{b}=0.2,l_r^{f}=0.05$. The feature map size is set as defined in Eqn.~\ref{eqn:hole1} and Eqn.~\ref{eqn:hole2}, the dimension of texture features is set as $16$, the dimension of feature embedding as $10$, and $\phi_\theta$ is a 4-layer MLP with hidden dimension $64$ and output dimension $3$. We utilize the Adam optimizer with decoupled learning rates to optimize all parameters for $1500$ epochs: the MLP parameters are optimized with an initial learning rate of $1e-3$, while the feature embeddings and texture features are set to $5e-3$. (The higher learning rate for feature embeddings facilitates faster capture of high-frequency texture details.) All learning rates are decayed by a factor of $0.8$ every $250$ epochs to ensure stable convergence. Furthermore, the hyperparameter $\gamma$ in the loss function is set to $0.001$ to balance the respective loss terms. On a single NVIDIA RTX 3090 GPU, the entire pipeline takes an average of $2-3$ minutes for a $512 \times 512$ natural image with a peak memory usage of 12 GB. When utilizing an NVIDIA A100 GPU, the optimization time is further reduced to approximately 1 minute. For animation-driven experiments, our method achieves a rendering rate of 4 frames per second (fps) with high geometric fidelity and realistic physical simulation.


\noindent\textbf{Benchmarks\&Metrics.} In this paper, we focus on evaluating the representation capability of complex textured natural images and the manipulation ability of important semantic components (objects) within the images. For image representation task, we randomly select $1000$ images from the ImageNet~\cite{deng2009imagenet} dataset for evaluation. In addition to commonly used metrics for measuring image reconstruction quality, such as PSNR, SSIM, and LPIPS, we also focus on comparing the representation efficiency of different methods. Specifically, we examine the representation time and parameter count required for a single image at the same resolution. For the image compression task, we conduct tests on the DIV2K~\cite{agustsson2017ntire} validation set and use standard metrics such as bits-per-pixel (bpp) and encoding/decoding speed (fps). For image editing tasks, we utilize OIR-bench and HumanEdit~\cite{bai2024humanedit} benchmark for our experiments. We follow prior works~\cite{fu2023guiding,sheynin2024emu,zhang2023magicbrush,yu2025anyedit} and adopt semantic similarity (\emph{i.e.}, CLIP-I and CLIP-T) and visual similarity (\emph{i.e.}, DINO similarity and L1 distance) to evaluate the effectiveness of the proposed method. For the image animation task, we focus on the consistency between the generated animation and the original image. Therefore, we use the FID score between the input image and the generated frames to evaluate this consistency. Additionally, we employ video generation metrics, Visual Quality (VQ), Temporal Consistency (TC), Dynamic Degree (DD) and Factual Consistency (FC) as defined in VideoScore~\cite{he2024videoscore}, to assess the quality of the generated animation.


\subsection{Image Representation and Compression}
\subsubsection{Comparison with Explicit Image Representations}
\begin{figure*}
\centering
\includegraphics[width=0.9\linewidth]{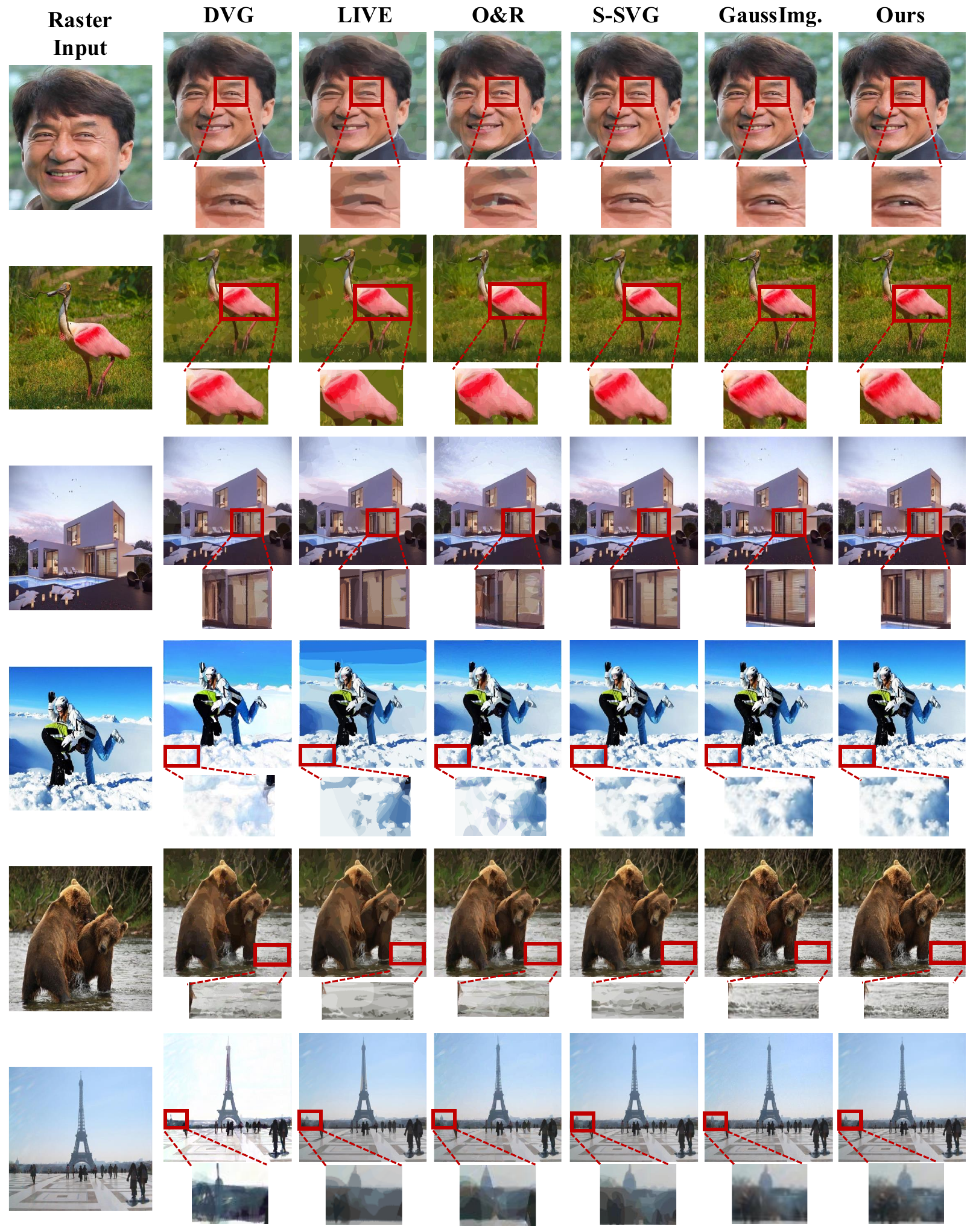}
\vspace{-0.2cm}
\caption{\textbf{Qualitative comparison on natural images.} We use red boxes to emphasize the differences. Our representation can express complex image details. The results of S-SVG~\cite{hu2024supersvg} are directly obtained from their paper. Please zoom in for more details.}
\label{fig:vis_imvec}
\end{figure*}

We compare our method with current SOTA explicit image representations (\emph{i.e.}, image vectorization methods (including DVG~\cite{Li:2020:DVG}, LIVE~\cite{ma2022towards}, O\&R~\cite{hirschorn2024optimize}, S-SVG~\cite{hu2024supersvg} and Chen~\emph{et al.}~\cite{chen2024towards}) and 2D Gaussian Splatting-based method GaussianImage~\cite{zhang2024gaussianimage}) on image representation task. For Emojis\&Icon datasets, we use \textit{MSE} distance as the metric. For Clipart and ImageNet benchmark, we use \textit{MSE}, \textit{PSNR}, \textit{LPIPS} and \textit{SSIM} as metrics. Note that S-SVG~\cite{hu2024supersvg} only releases part of its source code, thus we directly use the best results reported in their paper ($4000$ shapes). In addition, increasing the number of shapes only has a limited effect on the performance of GaussianImage~\cite{zhang2024gaussianimage} once the number reaches a certain threshold. Therefore, we use a fixed number of $10,000$ Gaussian kernels, resulting in a parameter count that is twice that of our method. For other compared image vectorization methods, we use $512$ primitives on Clipart and $4000$ primitives on ImageNet. Quantitative comparisons are shown in Tab.~\ref{tab:emojiicon} and Tab.~\ref{tab:cliimg}. We observe that our method consistently outperforms other methods across all metrics. In particular, existing methods are significantly short in natural image reconstruction compared to ours, which can be attributed to that previous approaches either stack shapes~\cite{Li:2020:DVG,ma2022towards,hu2024supersvg,hirschorn2024optimize} to represent textures or encode textures into sparse edge control points~\cite{chen2024towards}. In contrast, our method employs multi-layer multi-scale interrelated geometric control points to encode both global information and local texture details, achieving superior reconstruction quality.
We also show some qualitative comparisons in Fig.~\ref{fig:vis_imvec}. We observe that, while GaussianImage effectively captures sharp edge information, it is still limited by the stacking of irregular primitives, resulting in noticeable visual artifacts (\emph{e.g.}, the complex textures of the river surface and ground). Our method captures intricate and complex textures thanks to hierarchical texture embedding.

\begin{table}
\footnotesize
\caption{\textbf{Comparisons with image vectorization methods on Emoji\&Icon datasets.} \textit{MSE} ($\times10^{-3}$) results are reported. All compared methods use $10$ shapes. Our method achieves significantly better results.}
\begin{center}
\begin{tabular}{l|cccccc}
\toprule
Dataset&DVG&Im2Vec&LIVE&O\&R&Chen&\textbf{Ours} \\
\midrule
Emoji&9.24&25.81&1.61&1.44&0.63&\textbf{0.12} \\
Icon&28.52&32.90&2.42&2.23&0.70&\textbf{0.11} \\
\bottomrule
\end{tabular}
\end{center}
\label{tab:emojiicon}
\vspace{-0.2cm}
\end{table}

\begin{table}
\setlength{\tabcolsep}{4pt}
\footnotesize
\caption{\textbf{Comparisons with explicit parametric image representations on Clipart and ImageNet.} \textit{MSE} ($\times10^{-3}$) and other image reconstruction metrics are reported. Our method achieves best results across all metrics, especially on complex natural images. S-SVG is based on deep network training, thus we do not compare its running time. The time (\textit{mins}) is tested on an
NVIDIA GeForce RTX 3090 GPU.}
\begin{center}
\begin{tabular}{c|c|c|cccc}
    \toprule
    Dataset&Method&Time&MSE$\downarrow$&PSNR$\uparrow$&LPIPS$\downarrow$&SSIM$\uparrow$ \\
    \hline
    \multirow{5}{*}{Clipart.}
    &DVG&7.5&2.18&26.60&0.2998&0.8542 \\
    &LIVE&90.2&1.92&27.17&0.2633&0.8742 \\
    &O\&R&10.9&1.64&27.85&0.2599&0.8739 \\
    &GaussImg.&2.9&0.21&37.39&0.0883&0.9600 \\
&\textbf{Ours}&2.3&\textbf{0.17}&\textbf{37.83}&\textbf{0.0793}&\textbf{0.9610} \\
    \hline
    \multirow{6}{*}{ImageNet.}
    &DVG&89.8&2.63&25.80&0.3823&0.8459 \\
    &LIVE&\textgreater1000&2.91&25.36&0.4076&0.8443 \\
    &O\&R&81.3&2.98&25.26&0.4058&0.8464 \\
    &S-SVG&-&1.40&29.96&0.2496&0.9028 \\
    &GaussImg.&3.0&0.31&36.75&0.0857&\textbf{0.9617} \\
    &\textbf{Ours}&2.6&\textbf{0.30}&\textbf{36.99}&\textbf{0.0805}&0.9525 \\
    \bottomrule
\end{tabular}
\end{center}
\label{tab:cliimg}
\vspace{-0.2cm}
\end{table}

\subsubsection{Comparison with Implicit Image Representations}
\begin{figure}
\centering
\includegraphics[width=1.0\linewidth]{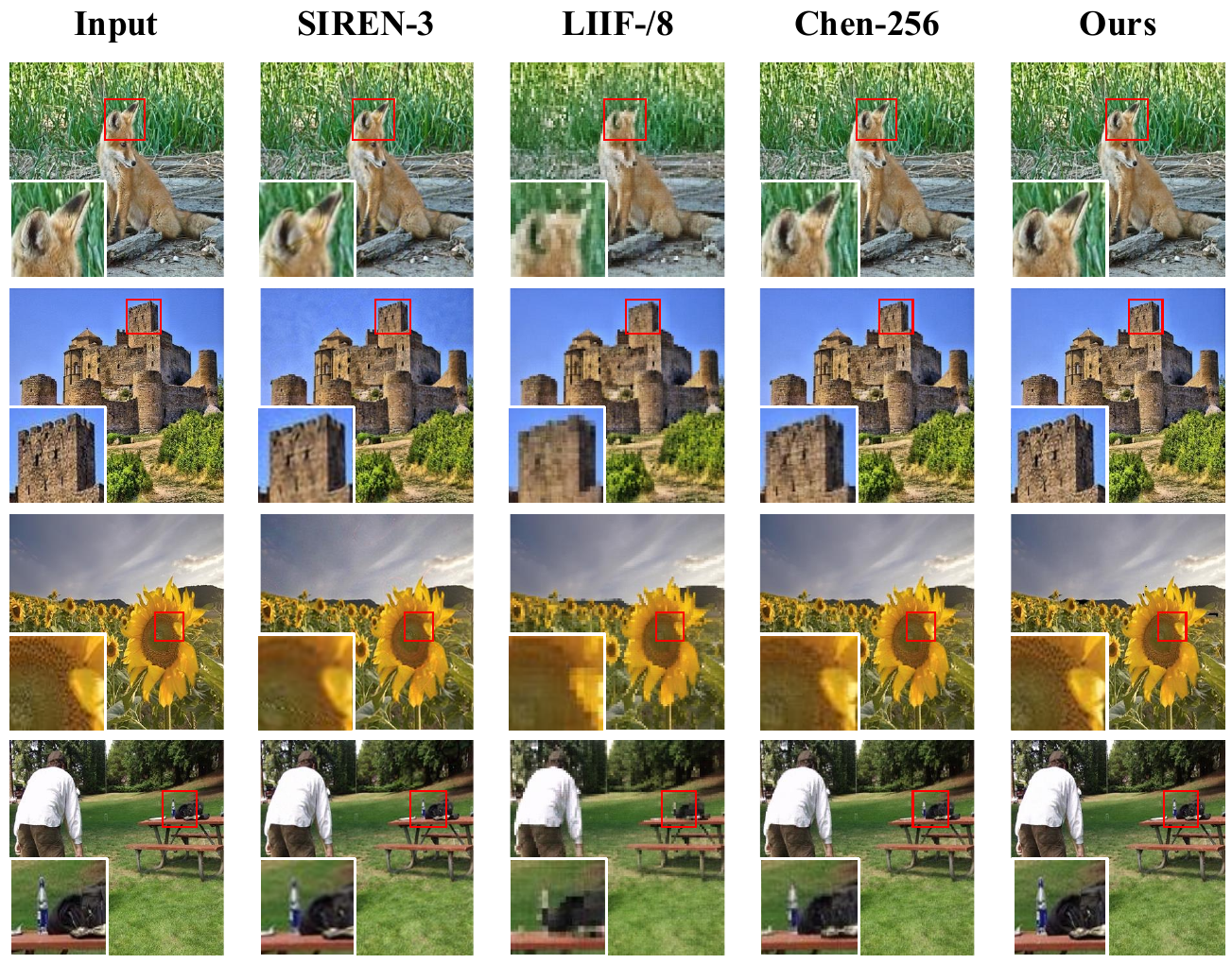}
\caption{\textbf{Qualitative comparison on natural images.} We visualize examples of methods with the same order of magnitude of parameters. Please zoom in for more details.}
\label{fig:vis_imp}
\end{figure}

We compare our method with implicit image representation algorithms (\emph{i.e.}, SIREN~\cite{sitzmann2020implicit}, Grid-based method and Chen~\emph{et al.}~\cite{chen2024towards}) on the ImageNet dataset. It is worth noting that although our method also employs grid-based feature planes, this serves only as an initialization strategy for node features. We distribute the features from the grid onto surrounding proxy nodes and then perform geometry-aware local feature interpolation. This process is fundamentally different from the grid-based approach used in the baseline method.
We use $3-$layer and $5-$layer version of SIREN for comparison. For grid-based method, we follow~\cite{chen2024towards} to utilize the framework of LIIF~\cite{chen2021learning} to perform zero-shot image reconstruction, which stores feature vectors on fixed grids and decodes the features with a $2-$layer MLP. We demonstrate the effectiveness and efficiency of our method by comparing reconstruction quality and the number of implicit parameters on the ImageNet dataset. Quantitative results are reported in Tab.~\ref{tab:imp}. The results indicate that our method achieves significantly higher reconstruction metrics than counterparts with a highly compressed parameter count. Notably, compared to Chen~\emph{et al.}~\cite{chen2024towards}, although we place many proxy geometric points within each shape, our layer-wise strategy to model each image layer as a single shape allows us to save a substantial number of B\'{e}zier control points (Chen~\emph{et al.}~\cite{chen2024towards} need several hundreds of $4-$segment shapes for a natural image while our method needs only a few shapes/image layers). In Fig.~\ref{fig:vis_imp}, we present a visual comparison with methods that have similar number of parameters to ours.
SIREN~\cite{sitzmann2020implicit} exhibits noticeable blurring and a lack of high-frequency details when the number of MLP layers is insufficient as it fits all information into MLP parameters, disregarding the explicit image structure and texture distribution. For grid-based method that stores features on fixed grids, a large number of features is needed to achieve acceptable reconstruction results, which is very inefficient. Chen~\emph{et al.}~\cite{chen2024towards} fail to model intricate texture details due to the sparse texture embedding.
Our method achieves high parameter compression while preserving high reconstruction quality by utilizing multi-layer and multi-scale texture embedding.

\begin{table}
\setlength{\tabcolsep}{4pt}
\footnotesize
\caption{\textbf{Comparison with general implicit image representations.} \textit{MSE} ($\times10^{-3}$) and other reconstruction results on ImageNet benchmark are reported. ``Codes'' denotes the number of feature vectors utilized. ``Params'' denotes the number of parameters. ``SIREN'' does not distribute feature vectors. ``SIREN-$\ast$'' means the SIREN version with $\ast$ layers. In ``LIIF-/$\ast$'', the ``$\ast$'' denotes the proportion of the spatial dimension of the original image to the grids storing feature vectors as defined in~\cite{chen2024towards}.}
\begin{center}
\begin{tabular}{l|cccccc}
\toprule
Method&MSE$\downarrow$&PSNR$\uparrow$&LPIPS$\downarrow$&SSIM$\uparrow$&Codes.$\downarrow$&Params.$\downarrow$ \\
\midrule
SIREN-3&2.50&26.02&0.2773&0.8267&-&50307 \\
SIREN-5&1.70&27.70&0.2605&0.8688&-&83331 \\
\midrule
LIIF-/8&6.58&21.82&0.5059&0.6347&4096&68099 \\
LIIF-/4&3.10&25.09&0.2920&0.8241&16384&264707 \\
\midrule
Chen-256&1.64&27.85&0.2595&0.8692&3072&58115 \\
Chen-512&1.56&28.07&0.2593&0.8717&6144&113411 \\
\midrule
\textbf{Ours}&\textbf{0.30}&\textbf{36.99}&\textbf{0.0805}&\textbf{0.9525}& \textbf{263} & 59955 \\
\bottomrule
\end{tabular}
\end{center}
\label{tab:imp}
\end{table}

\subsubsection{Comparison with Image Compression Methods}
We also make comparisons on the image compression task with state-of-the-art methods. In our representation, the majority of parameters originate from the coordinates and texture feature vectors of the proxy nodes, as well as from the decoding function (MLP $\phi_\theta$). To enable a fair comparison with image compression methods, we quantize the parameters used in our representation. Specifically, we apply 8-bit quantization directly to the coordinates of the proxy nodes. For the texture feature vectors (feature planes) on the proxy nodes and the decoding function, any standard quantization techniques~\cite{li2021brecq,guo2022squant,lin2023bit} for neural networks can be readily applied. In practice, we employ the BRECQ~\cite{li2021brecq} for 4-bit quantization. The results on the DIV2K~\cite{agustsson2017ntire} validation set are shown in Tab.~\ref{tab:compression}. Please note that the performance metrics of the compared methods in the table are derived from the source paper or GaussianImage~\cite{zhang2024gaussianimage}. To ensure a fair comparison with previous works, we also evaluate the efficiency-related metrics on an NVIDIA V100 GPU. As we observe, our method, even after quantization, still achieves comparable image reconstruction results and encoding/decoding speeds. The speed difference arises from the rasterization-vs-inference nature of the methods. While GaussImg~\cite{zhang2024gaussianimage} leverages established graphics pipelines, our method benefits from the growing efficiency of neural accelerators and future optimizations like quantization. In addition, all the comparison methods are designed solely for image compression and do not support other image manipulation tasks. In contrast, our representation not only delivers strong image compression performance but also supports efficient and controllable image editing and animation.

\begin{table}
\setlength{\tabcolsep}{4pt} 
\centering
\caption{Comparison with \textbf{image compression} methods on the \textbf{DIV2K} validation set.}
\begin{tabular}{l|ccccc}
\toprule
Method & Bpp$\downarrow$ & PSNR$\uparrow$ & SSIM$\uparrow$ & Enc. FPS$\uparrow$ & Dec. FPS$\uparrow$ \\
\midrule
JPEG~\cite{wallace1991jpeg} & 0.5638 & 28.43 & 0.9559 & 557.35 & 545.59 \\
JPEG2000~\cite{skodras2002jpeg} & 0.5993 & 30.93 & 0.9663 & 3.40 & 3.93 \\
Ballé17~\cite{balle2016end} & 0.4987 & 30.78 & 0.9775 & 16.53 & 17.87 \\
Ballé18~\cite{balle2018variational} & 0.5415 & 32.23 & 0.9816 & 13.56 & 15.20 \\
COIN~\cite{dupont2021coin} & 0.6780 & 27.61 & 0.9306 & $3.51\times10^{-4}$ &93.74 \\
GaussImg.~\cite{zhang2024gaussianimage} & 0.6417 & 27.56 & 0.9483 & $4.73\times10^{-3}$ & 1980.54 \\
\textbf{Ours} & 0.6409 & 27.13 & 0.9298 & $5.13\times10^{-3}$ & 4.02  \\
\bottomrule
\end{tabular}
\label{tab:compression}
\vspace{-0.2cm}
\end{table}

\subsection{Image Manipulation}
\subsubsection{Image Geometry Editing}
In the image geometry editing experiments, we first conduct quantitative evaluations on the HumanEdit~\cite{bai2024humanedit} benchmark, comparing our method with advanced open-source approaches. In addition, we perform qualitative comparisons against state-of-the-art commercial models, including Gemini 2.5 Flash Image (Nano Banana)~\cite{comanici2025gemini} and GPT-Image\cite{GPT_Image}, to further demonstrate the effectiveness of our approach. The editing tasks encompass the following types: \emph{Add, Counting, Remove, and Relation}, which cover the target editing types of our method. The quantitative results are shown in Tab.~\ref{tab:humanedit_results}. We observe that our method outperforms classical generative image editing models on nearly all metrics. Notably, in terms of the L1, L2, CLIP-I, and DINO metrics, our method achieves performance far surpassing previous approaches across all editing tasks. This is due to the effective decoupling of image semantics, geometry, and texture enabled by our parametric proxy embedding, which allows for precise editing of specific attributes of targeted objects without affecting other regions or attributes.

\begin{table}
\centering
\caption{Quantitative comparisons of \textbf{geometric editing} task on \textbf{HumanEdit} benchmark.}
\label{tab:humanedit_results}
\begin{tabular}{lccccc}
\toprule
Method & L1$\downarrow$ & L2$\downarrow$ & CLIP-I$\uparrow$ & DINO$\uparrow$ & CLIP-T$\uparrow$ \\
\midrule
\multicolumn{6}{c}{\textbf{HumanEdit-Add}} \\
\midrule
InstructPix2Pix  & 0.1152 & 0.0329 & 0.8135 & 0.6230 & 0.2764 \\

HIVE  & 0.0885 & 0.0234 & 0.8863 & 0.7811 & 0.2706 \\
MagicBrush  & 0.0580 & 0.0167 & 0.9102 & 0.8562 & 0.2745 \\

AnySD & 0.0870 & 0.0307 & 0.9137 & 0.8379 & 0.3047 \\
UltraEdit & 0.0805 & 0.0216 & 0.9158 & 0.8112 & 0.2956 \\
\textbf{Ours}  & \textbf{0.0255} & \textbf{0.0058} & \textbf{0.9598} & \textbf{0.9105} & \textbf{0.3106} \\

\midrule
\multicolumn{6}{c}{\textbf{HumanEdit-Counting}} \\
\midrule
InstructPix2Pix  & 0.1628 & 0.0586 & 0.8124 & 0.5850 & 0.2716 \\

HIVE  & 0.1211 & 0.0442 & 0.8826 & 0.7431 & 0.2705 \\
MagicBrush  & 0.1058 & 0.0434 & 0.8677 & 0.7103 & 0.2707 \\
AnySD & 0.0362 & 0.0071 & 0.9715 & 0.9273 & 0.2709 \\
UltraEdit & 0.0561 & 0.0162 & 0.9266 & 0.8502 & 0.2665 \\
\textbf{Ours}  & \textbf{0.0157} & \textbf{0.0025} & \textbf{0.9738} & \textbf{0.9640} & \textbf{0.2723} \\
\midrule
\multicolumn{6}{c}{\textbf{HumanEdit-Remove}} \\
\midrule
InstructPix2Pix  & 0.1624 & 0.0504 & 0.7240 & 0.4188 & 0.2325 \\

HIVE  & 0.1179 & 0.0375 & 0.8362 & 0.6562 & 0.2474 \\
MagicBrush  & 0.0690 & 0.0232 & 0.8985 & 0.8249 & 0.2572 \\
AnySD & 0.1029 & 0.0451 & 0.9366 & 0.8543 & 0.2617 \\
UltraEdit & 0.0683 & 0.0194 & 0.9120 & 0.8632 & 0.2724 \\
\textbf{Ours}  & \textbf{0.0166} & \textbf{0.0033} & \textbf{0.9659} & \textbf{0.9785} & \textbf{0.2739} \\
\midrule
\multicolumn{6}{c}{\textbf{HumanEdit-Relation}} \\
\midrule
InstructPix2Pix  & 0.1741 & 0.0647 & 0.8069 & 0.5851 & 0.2828 \\

HIVE  & 0.1298 & 0.0460 & 0.8689 & 0.7005 & 0.2793 \\
MagicBrush  & 0.0884 & 0.0334 & 0.8985 & 0.7865 & 0.2823 \\
AnySD & 0.0722 & 0.0305 & 0.9357 & 0.8847 & 0.2822 \\
UltraEdit & 0.0633 & 0.0229 & 0.9057 & 0.8805 & 0.2926 \\
\textbf{Ours}  & \textbf{0.0173} & \textbf{0.0051} & \textbf{0.9604} & \textbf{0.9552} & \textbf{0.2957} \\

\bottomrule
\end{tabular}%
\vspace{-0.2cm}
\end{table}

\begin{figure*}
\centering
\includegraphics[width=0.95\linewidth]{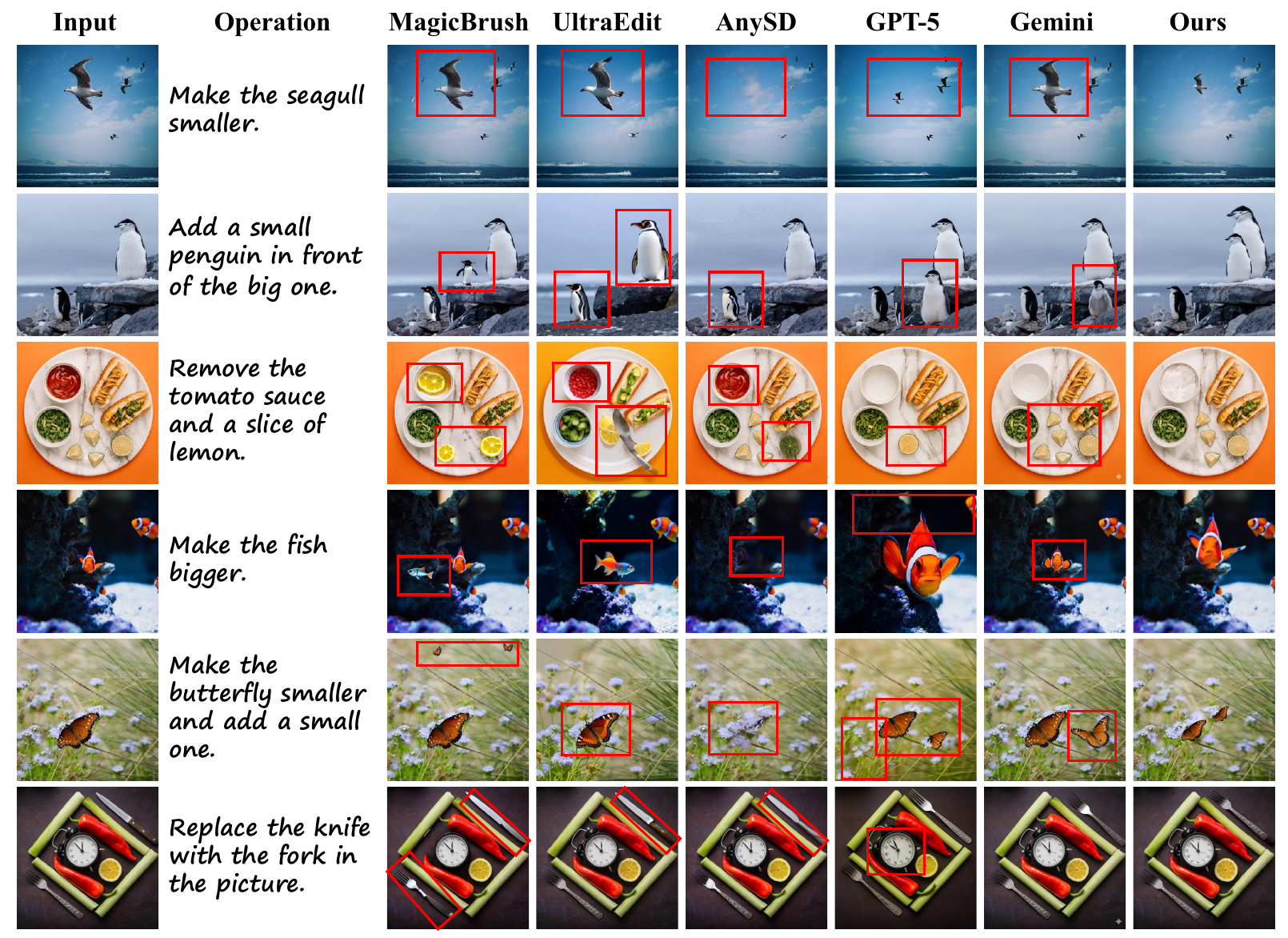}
\caption{\textbf{Qualitative comparisons of Geometric Editing task.} Red boxes highlight the uncontrollability in competing methods, demonstrating their limited ability to perform fine-grained, high-fidelity geometric edits. While other models struggle to precisely modify targets without affecting the background, our method produces clean, instruction-faithful results. Note that ``Gemini'' means Gemini 2.5 Flash Image~\cite{comanici2025gemini}. Please zoom in for further details.}
\label{fig:geo_vis}
\end{figure*}

We compare our method with approaches that achieve strong quantitative performance, as well as several state-of-the-art commercial models, and report the qualitative results as shown in the figure. We observe that image-editing–specific models often fail to align their outputs with the given instructions, largely due to their limited instruction understanding and the entangled structures within their latent spaces. Even advanced commercial systems exhibit notable limitations. For example, Nano-banana performs a certain degree of spatial disentanglement during editing, yet it still fails to respond to geometric editing commands (\emph{e.g.}, the 1st and 4th rows of the 7th column). GPT-Image can follow geometric instructions to some extent, but it inevitably introduces severe artifacts in non-edited regions (\emph{e.g.}, the 3rd, 4th, and 6th rows of the 6th column), revealing its lack of controllability. It is also worth noting that existing generative models exhibit substantial randomness during editing, often requiring users to repeatedly adjust prompts until a reasonable result can be obtained. In contrast, our method remains strictly aligned with user-specified instructions and produces precise, controllable, and stable edits with high efficiency.

\subsubsection{Image Texture Editing}
We use the \emph{Replace} editing type from HumanEdit~\cite{bai2024humanedit} and the whole OIR-Bench~\cite{yang2023object} to test the texture editing capabilities. Given that existing comparison methods (including InstructPix2Pix~\cite{brooks2023instructpix2pix}, MagicBrush~\cite{zhang2023magicbrush}, HIVE~\cite{zhang2024hive}, UltraEdit~\cite{zhao2024ultraedit} and AnyEdit~\cite{yu2025anyedit}) rely on instruction-based generative models for texture editing, we present only the quantitative results for generative texture editing, while the results for interactive texture editing are showcased through qualitative visualizations. The quantitative results are presented in Tab.~\ref{tab:humanedit_texture} and Tab.~\ref{tab:oirbench}. It is worth noting that OIR-Bench does not provide edited ground-truth images. Therefore, when computing quantitative metrics, in addition to evaluating the alignment between the editing results and the textual instructions (CLIP-T), we also measure the similarity between the edited image and the original input (\emph{i.e.}, SSIM and LPIPS), which reflects the extent to which the editing process alters the original content. As shown in the tables, our method consistently outperforms existing open-source models across all metrics on both benchmarks. In particular, we observe substantial improvements in similarity-based metrics such as L1, L2, MS-SSIM, and LPIPS. This indicates that, although our generative texture editing relies on a pre-trained generative model and Score Distillation Sampling (SDS), the proposed proxy representation effectively constrains the SDS optimization to modify only the texture of the target instance, without affecting other instances or their attributes. Consequently, our approach enables precise and controllable texture edits. 
\begin{table}
\centering  
\caption{Quantitative comparisons of \textbf{texture editing} task on \textbf{OIR} benchmark.}  
\begin{tabular}{lccc}  
\toprule  
Method & CLIP-T$\uparrow$ & MS-SSIM$\uparrow$ & LPIPS$\downarrow$ \\
\midrule  
InstructPix2Pix & 0.278 & 0.803 & 0.380 \\
HIVE & 0.284 & 0.729 & 0.395 \\
MagicBrush & 0.300 & 0.748 & 0.252  \\
AnySD & 0.270 & 0.785 & 0.352  \\
UltraEdit & 0.296 & 0.731 & 0.295 \\
\textbf{Ours} & \textbf{0.322} & \textbf{0.824} & \textbf{0.250} \\
\bottomrule  
\end{tabular}
\label{tab:oirbench} 
\vspace{-0.2cm}
\end{table}

\begin{table}
\centering
\caption{Quantitative comparisons of \textbf{texture editing} task on \textbf{HumanEdit} benchmark.}
\label{tab:humanedit_texture}
\begin{tabular}{lccccc}
\toprule
Method & L1$\downarrow$ & L2$\downarrow$ & CLIP-I$\uparrow$ & DINO$\uparrow$ & CLIP-T$\uparrow$ \\

\midrule
InstructPix2Pix  & 0.1910 & 0.0770 & 0.7887 & 0.5692 & 0.2697 \\
HIVE  & 0.1265 & 0.0443 & 0.8582 & 0.7087 & 0.2726 \\
MagicBrush  & 0.0984 & 0.0409 & 0.8757 & 0.7513 & 0.2716 \\
AnySD & 0.0824 & 0.0269 & 0.9367 & 0.8190 & 0.2730 \\
UltraEdit & 0.0362 & 0.0117 & 0.9398 & 0.8543 & 0.2673 \\
\textbf{Ours}  & \textbf{0.0332} & \textbf{0.0103} & \textbf{0.9461} & \textbf{0.8629} & \textbf{0.2772} \\
\bottomrule
\end{tabular}%
\vspace{-0.2cm}
\end{table}

\begin{figure*}
\centering
\includegraphics[width=0.95\linewidth]{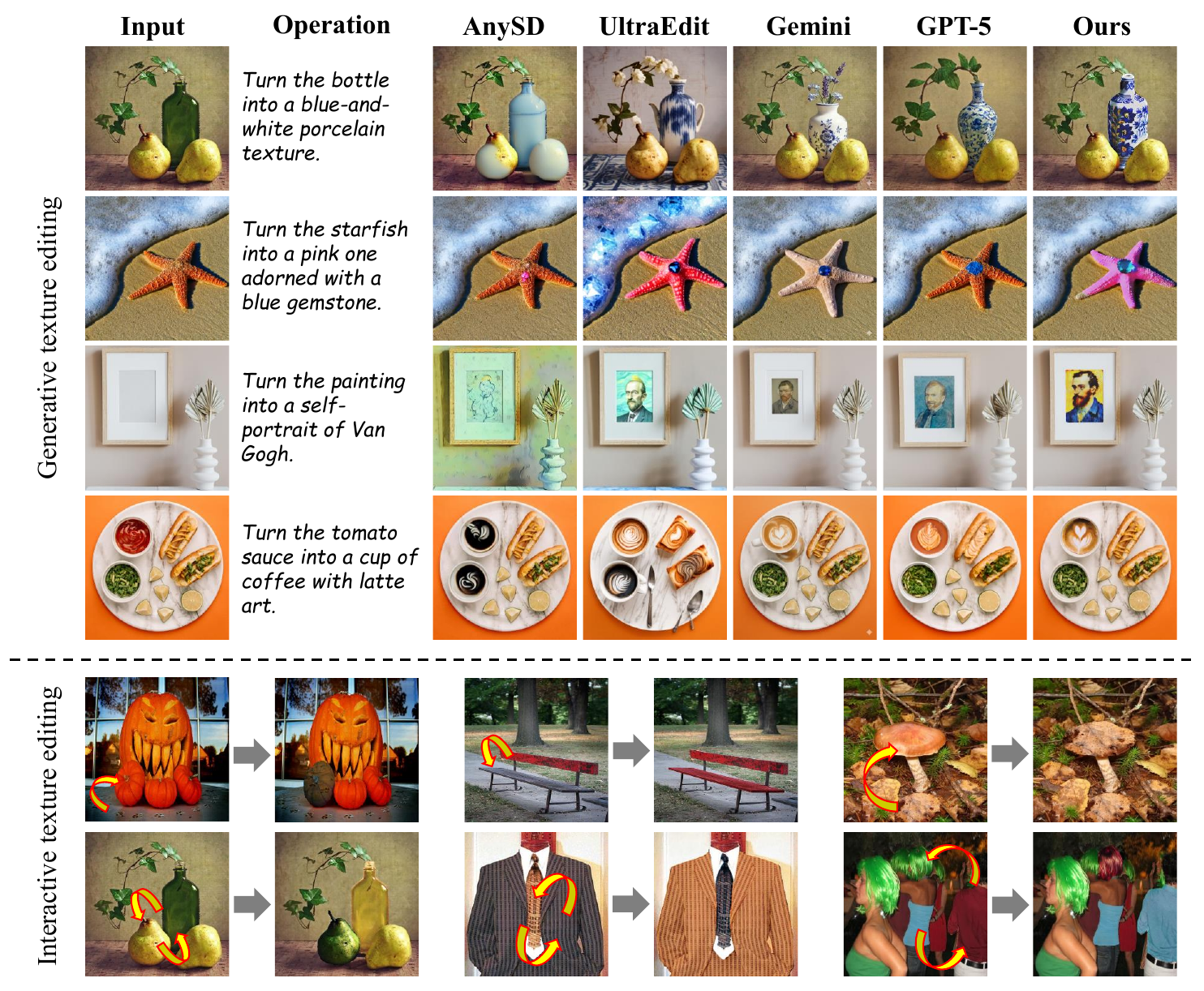}
\caption{\textbf{(Top): Qualitative Comparisons of the Generative Texture Editing Task.} Our method ensures absolute control over instance-level semantics, geometry, and other attributes during texture editing. (Gemini is powered by Nano Banana.)
\textbf{(Bottom): Qualitative Results of the Interactive Texture Editing Task.} We demonstrate the convenience and controllability of texture transfer within images. The yellow arrow represents the mouse operation (the direction of texture transfer).}
\label{fig:tex_vis}
\vspace{-0.3cm}
\end{figure*}

We compare our approach with state-of-the-art methods that perform well on the benchmark, as well as advanced commercial models, with visual results shown in Fig.~\ref{fig:tex_vis}. As illustrated, even cutting-edge commercial systems such as GPT-Image fail to preserve the geometric structure of the target instance during instance-level texture editing. In contrast, our method achieves strictly controllable instance-level texture manipulation, owing to our disentangled representation of geometry and texture. During the editing process, we modify only the texture parameters (feature planes) while keeping the geometric parameters (control point positions) unchanged. This design ensures that texture edits do not inadvertently alter the underlying instance geometry. 

Additionally, we present the results of our interactive texture editing pipeline, as shown in Fig.~\ref{fig:tex_vis}. As demonstrated, users can achieve instance-level texture modification simply by manually swapping the feature planes between different instances, while preserving both the geometric structure and semantic attributes of each instance. This highlights the strong potential of our approach for controllable and interactive texture editing applications.


\subsubsection{Image Animation}
\begin{figure*}
\centering
\includegraphics[width=1\linewidth]{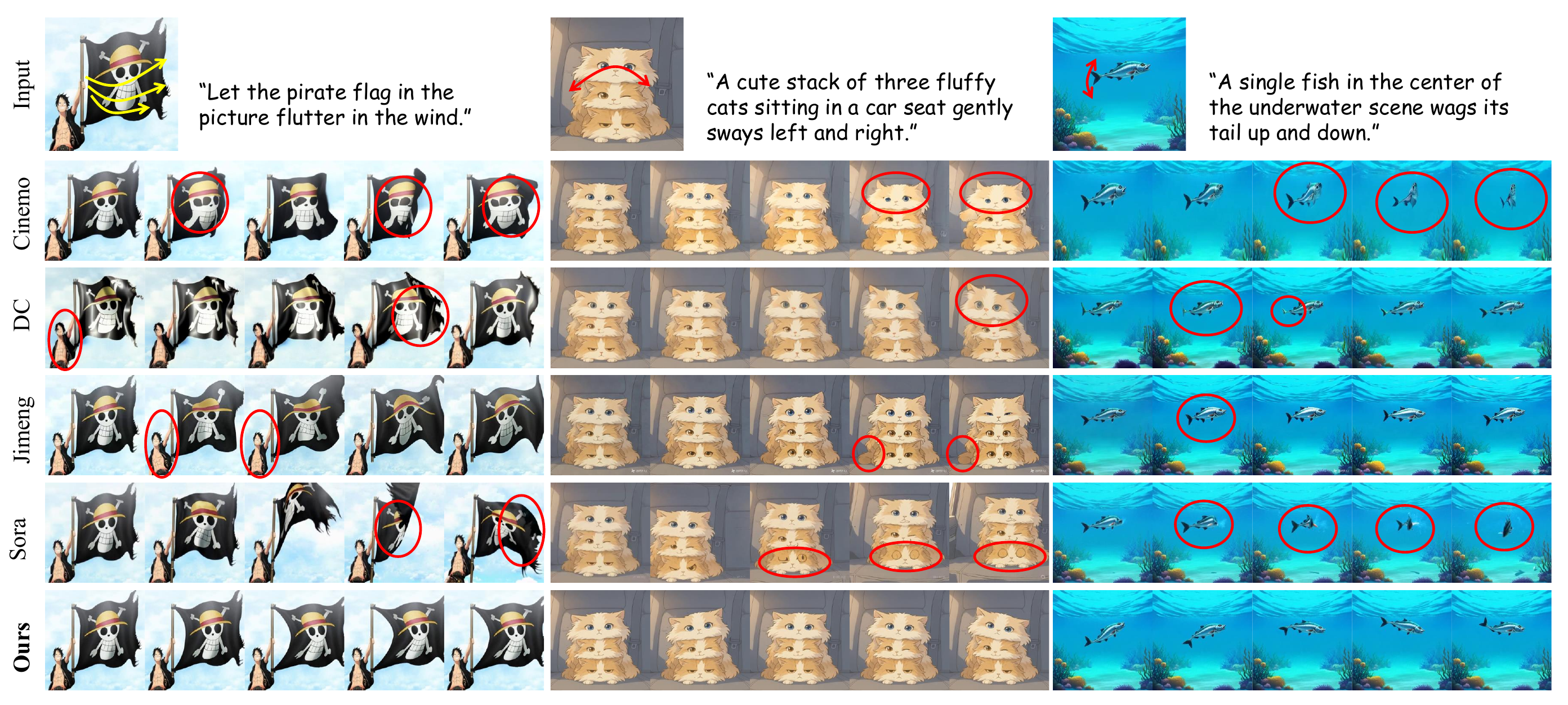}
\caption{\textbf{Qualitative Comparisons of the Image Animation Task.} Current state-of-the-art open-source methods fail to achieve basic temporal consistency. Even advanced commercial models, such as Sora, struggle to balance physical realism with dynamic fidelity. Our method achieves physically realistic, temporally consistent image-to-animation generation. (Note: Jimeng showen here is powered by Seedance 1.0 model~\cite{gao2025seedance}.)}
\label{fig:ani_vis}
\vspace{-0.3cm}
\end{figure*}

Given the absence of established benchmarks for animestyle image-to-video generation, we follow prior works~\cite{shi2024motion,xie2025physanimator} to construct a test set comprising 20 anime images with stylized dynamic elements such as fire, clothing, and mollusks. In the image animation task, we compare our method with classic open-source approaches, including Cinemo~\cite{ma2024cinemo}, DynamiCrafter~\cite{xing2024dynamicrafter}, and Motion-I2V~\cite{shi2024motion}, as well as popular commercial models such as Jimeng~\cite{jimeng} (powered by Seedance 1.0~\cite{gao2025seedance}) and Sora~\cite{sora}. We use video generation metrics mentioned above to evaluate the quality of the generated animations. The results are shown in Tab.~\ref{tab:image_anime}. Our method surpasses open-source baselines in visual quality, temporal consistency, and factual consistency, demonstrating overall superior video quality, despite a marginally lower video dynamic degree score compared to Cinemo and Motion-I2V, whose higher scores are accompanied by artifactual motions that compromise realism. Results are shown in Fig.~\ref{fig:ani_vis}.

\begin{table}
\centering  
\caption{Quantitative comparisons on \textbf{image animation} task.}  
\begin{tabular}{lccccc}  
\toprule  
Method & FID$\downarrow$ & VSVQ$\uparrow$ & VSTC$\uparrow$ & VSDD$\uparrow$ & VSFC$\uparrow$ \\
\midrule  
Cinemo~\cite{ma2024cinemo} & 92.3 & 2.853 & 2.439 & \textbf{2.828} & 2.351 \\
DynamiCrafter~\cite{xing2024dynamicrafter} & 87.2 & 2.875 & 2.625 & 2.684 & 2.528 \\
Motion-I2V~\cite{shi2024motion} & 116.9 & 2.809 & 2.516 & 2.816 & 2.368 \\
\textbf{Ours} & \textbf{52.5} & \textbf{3.056} & \textbf{2.704} & 2.711 & \textbf{2.587} \\
\bottomrule  
\end{tabular}
\label{tab:image_anime} 
\end{table}

\subsection{Ablation Study}
In this section, we perform ablated experiments to explore the important components and hyper-parameters of our method. For simplicity, all the quantitative results in this section are conducted on the image representation task.

\noindent\textbf{Component Analyses.} We first conduct experiments to explore the effectiveness of the crucial modules in the proposed framework. The results are shown in Tab.~\ref{tab:ablate}. We can see that when using only edge control points without internal triangles (``Edge-only''), reconstruction results on natural images deteriorate significantly. Employing a single level of internal triangles within shapes yields competitive results with relatively few parameters. Utilizing multi-scale triangles further enhances reconstruction quality with only a minimal increase in parameters, which demonstrates the effectiveness of our multi-scale scheme. We also conduct the ablated version without B\'{e}zier fitting, where we directly perform Triwild algorithm on the entire image (``Tri-only''). With a similar parameter count, this version exhibits reduced reconstruction quality on natural images. This demonstrates the effectiveness of our layer-wise strategy and multi-level feature fusion at both shape edges and interiors.
We also demonstrate the effectiveness of the feature encoding and the interpolation method based on barycentric coordinates, which further enhance our method.

\begin{figure}
\centering
\includegraphics[width=1\linewidth]{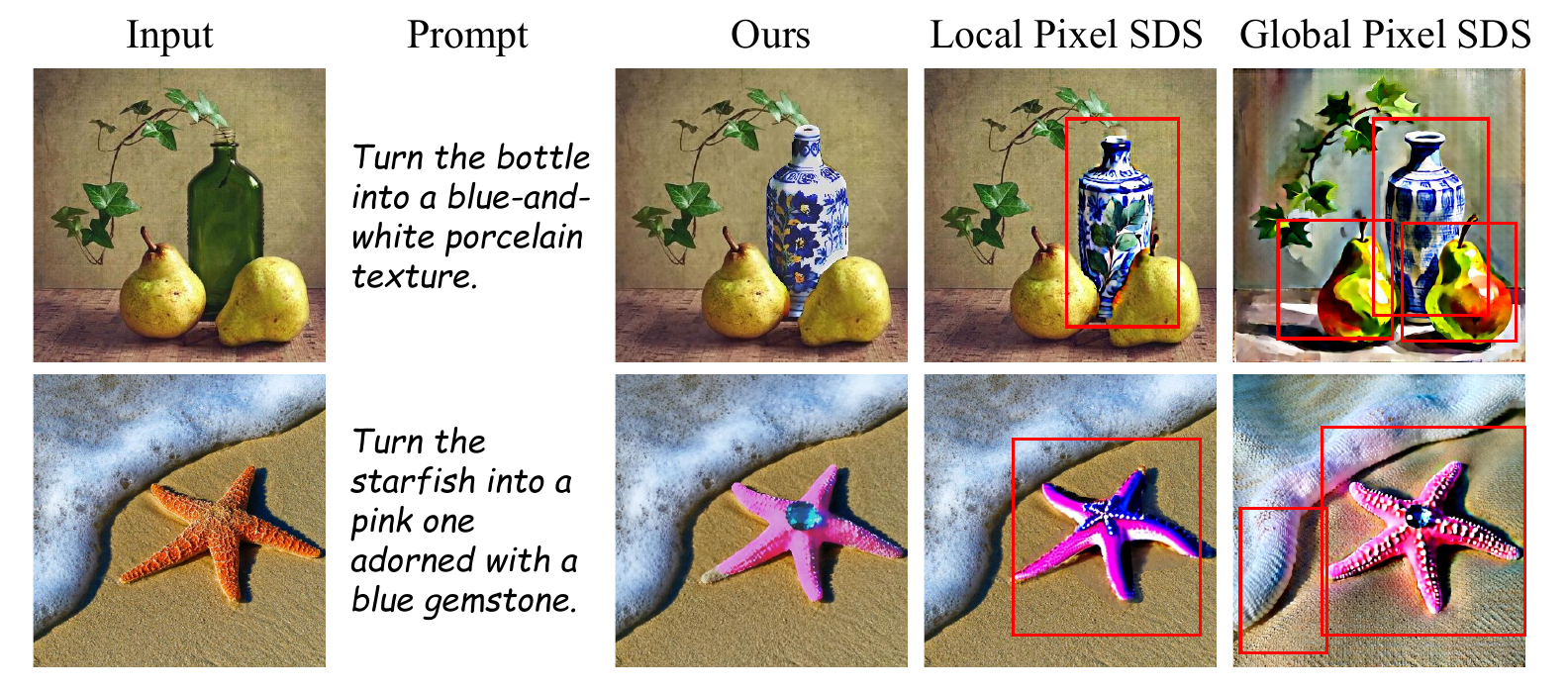}
\vspace{-0.6cm}
\caption{Ablation study on Proxy Embedding versus Pixel Representation.}
\label{fig:tex_ablation}
\vspace{-0.2cm}
\end{figure}

We also conducted an ablation study to demonstrate the effectiveness of the proxy embedding. Qualitative results are shown in Fig.~\ref{fig:tex_ablation}. We can observe that, when editing pixels solely using the SDS-based algorithm—whether editing all pixels or only those within an instance—competitive texture editing results cannot be achieved. In contrast, when using the proposed proxy embedding, SDS optimization is applied only to the embedding of the target instance. This not only ensures instance-level decoupling but also reduces the optimization complexity, enabling controllable and realistic texture editing. 

\begin{figure}
\centering
\includegraphics[width=1\linewidth]{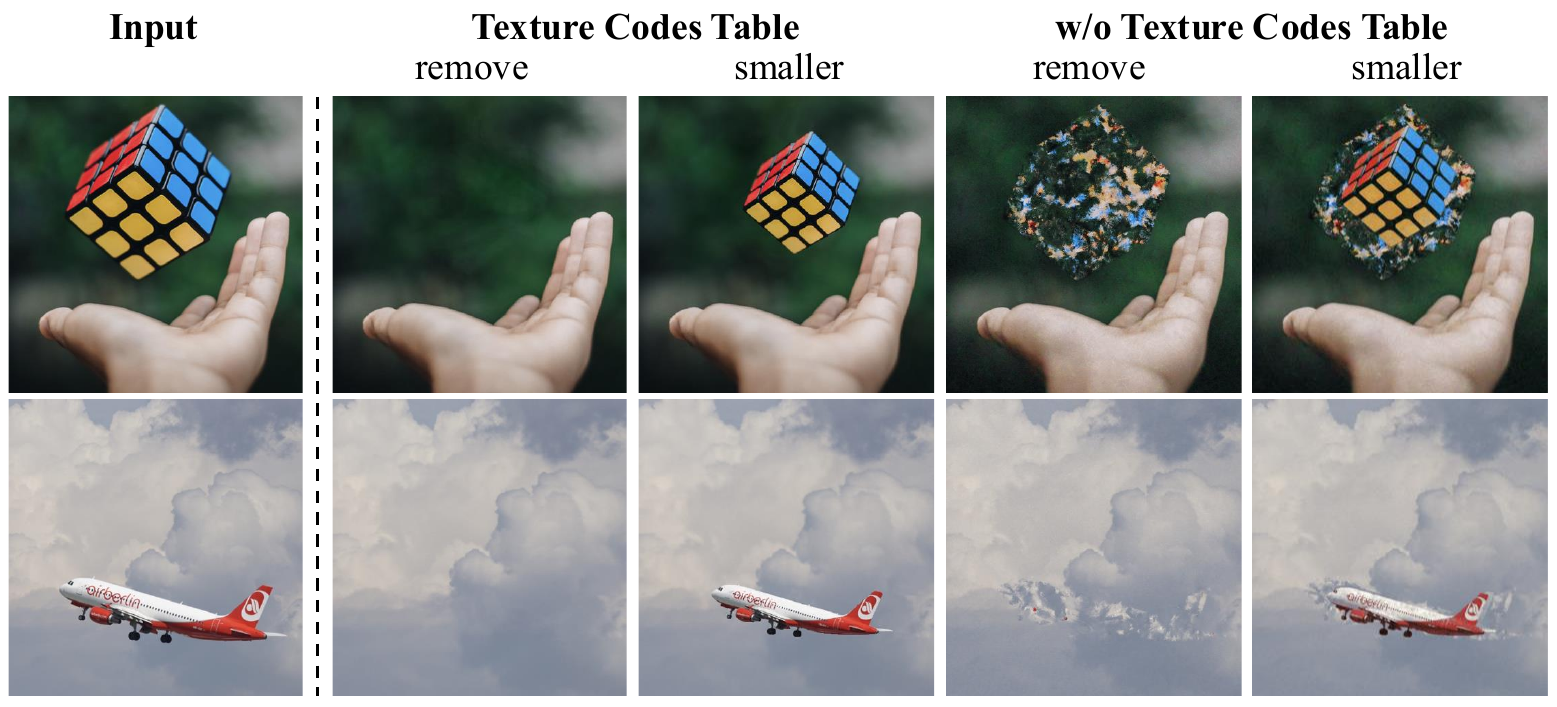}
\vspace{-0.5cm}
\caption{Analyses on the impact of Locality-Sensitive Texture Codes Table.}
\label{fig:bg_ablation}
\vspace{-0.2cm}
\end{figure}

Additionally, we conducted an ablation study to demonstrate the impact of using the Locality-Sensitive Texture Codes Table compared to a random initialization and independent optimization approach for all proxy nodes on the editing performance. The qualitative results, as shown in Fig.~\ref{fig:bg_ablation}, highlight that the proposed Locality-Sensitive Texture Codes Table successfully associates local texture features of patterns in the image. This enables seamless and natural edge texture continuity in hole completion, laying a crucial foundation for subsequent image editing and animation tasks.

\begin{table}
\setlength{\tabcolsep}{4pt} 
\caption{\textbf{Component analyses on feature embedding methods.} ``SinTri'' and ``MulTri'' denotes single-scale triangles and multi-scale triangles respectively. ``IDW-Tri'' means using inverse distance weighting interpolation within triangles. \textit{MSE} ($\times10^{-3}$) results and parameter counts on ImageNet are reported. ``Random Init.'' refers to randomly initializing and independently optimizing the texture vectors on all proxy nodes, without using a unified feature plane for allocation.}
\footnotesize
\begin{center}
\begin{tabular}{c|ccccccc}
\toprule
\multirow{2}{*}{~}&Edge&Edge+&Edge+&Tri&w/o Feat&IDW&Random \\
~&-only&SinTri&MulTri&-only&Enc.&-Tri&Init. \\
\midrule
MSE&1.02&0.49&\textbf{0.30}&0.57&0.88&0.39&0.34 \\
Params.&59955&59955&59955&59955&39216&59955&89120 \\
\bottomrule
\end{tabular}  
\end{center}
\label{table:ablate1}
    \label{tab:ablate}
\vspace{-0.5cm}
\end{table}

\begin{figure}[H]
\centering
\includegraphics[width=1.0\linewidth]{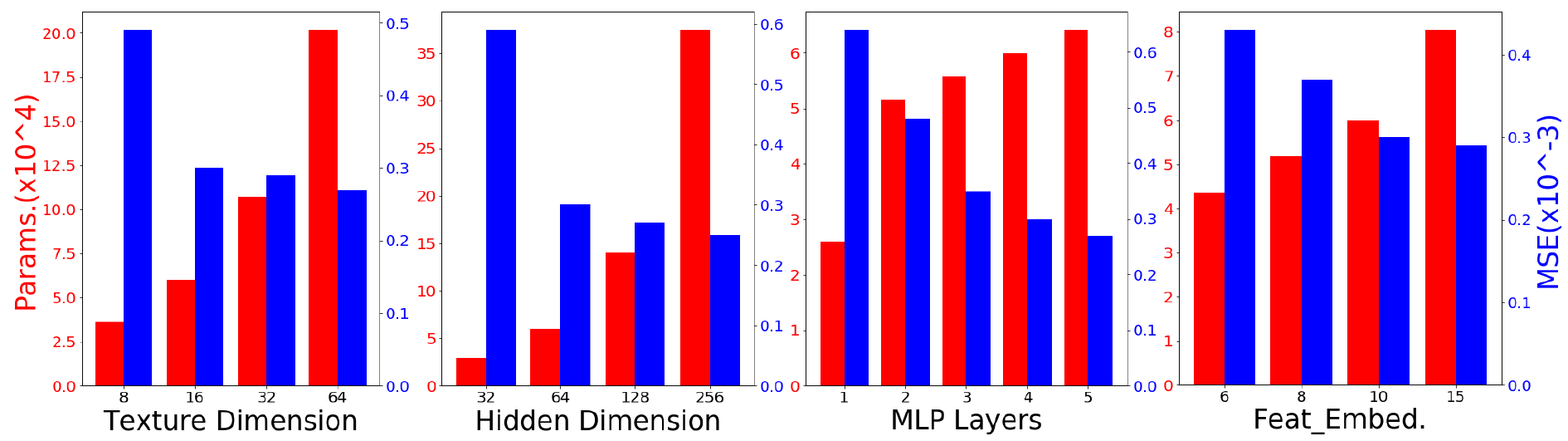}
\vspace{-0.5cm}
\caption{\textbf{Parameter analyses.} \textit{MSE} ($\times10^{-3}$) results (blue) and parameter counts (red) on ImageNet are reported.}
\label{fig:ablation2}
\end{figure}

\noindent\textbf{Parameter Analyses.} We investigate how the texture feature dimension, the hidden dimension and the number of MLP layers and the texture feature embedding frequency affect the reconstruction results. The results are shown in Fig.~\ref{fig:ablation2}. We observe that, due to our efficient multi-layer multi-scale texture feature embedding, the improvement in reconstruction quality from higher texture feature dimensions or higher hidden dimensions is minimal. Additional MLP layers further enhance the performance of our method. Considering the trade-off between reconstruction quality and parameter counts, using a four-layer MLP proves to be highly efficient.

\section{Conclusion}
This work introduces ProxyImg, a novel and efficient image representation that parameterizes any image into layer-wise texture embeddings distributed across hierarchical proxy nodes. Extensive experiments across a range of tasks and benchmarks demonstrate that our representation not only reconstructs complex natural images with substantial parameter compression but also enables precise, controllable, and multi-turn image editing with minimal effort. This high level of controllability allows for any desired image modification, ensuring flexibility and ease of editing. Furthermore, ProxyImg facilitates the generation of physically realistic and temporally consistent image-to-animation sequences, offering a powerful tool for efficient and accurate image-based animation. These results highlight the potential of ProxyImg for a wide range of image editing and generation applications, showcasing its effectiveness in both practical and creative contexts.





\bibliographystyle{IEEEtran}
\bibliography{main}
%

 

\vspace{-0.5cm}
\begin{IEEEbiography}[{\includegraphics[width=1in,height=1.25in,clip,keepaspectratio]{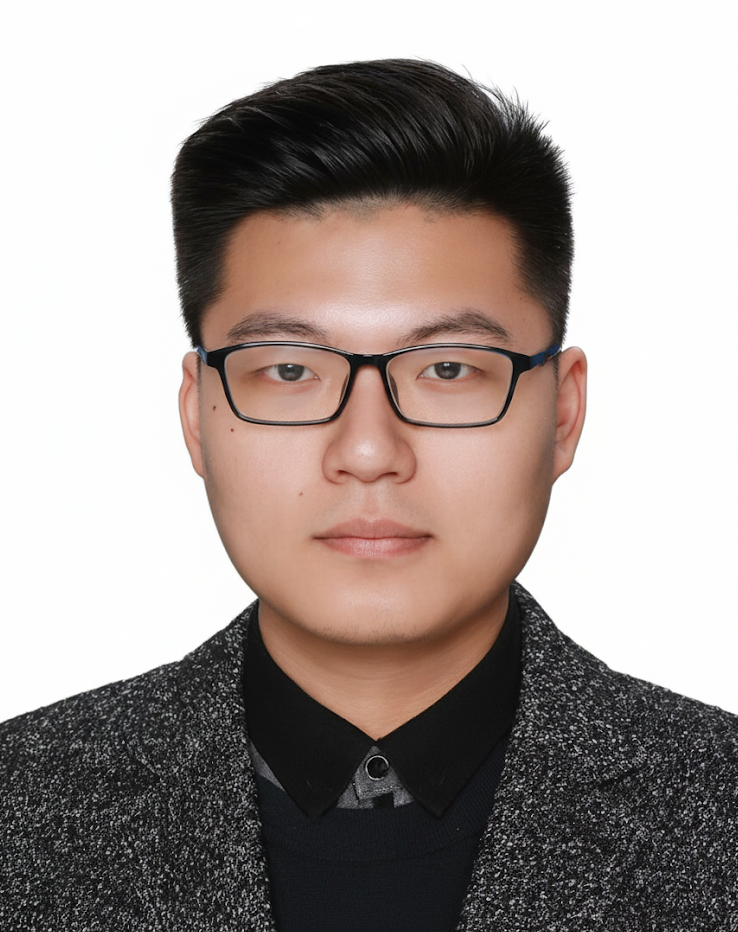}}]{Ye Chen} received his Bachelor's and Master's degrees in Electronic Engineering from Shanghai Jiao Tong University in 2020 and 2023, respectively. He is currently working toward his Ph.D. degree at Shanghai Jiao Tong University. His research interests include controllable visual generation and physics-based simulation.
\end{IEEEbiography}

\begin{IEEEbiography}[{\includegraphics[width=1in,height=1.25in,clip,keepaspectratio]{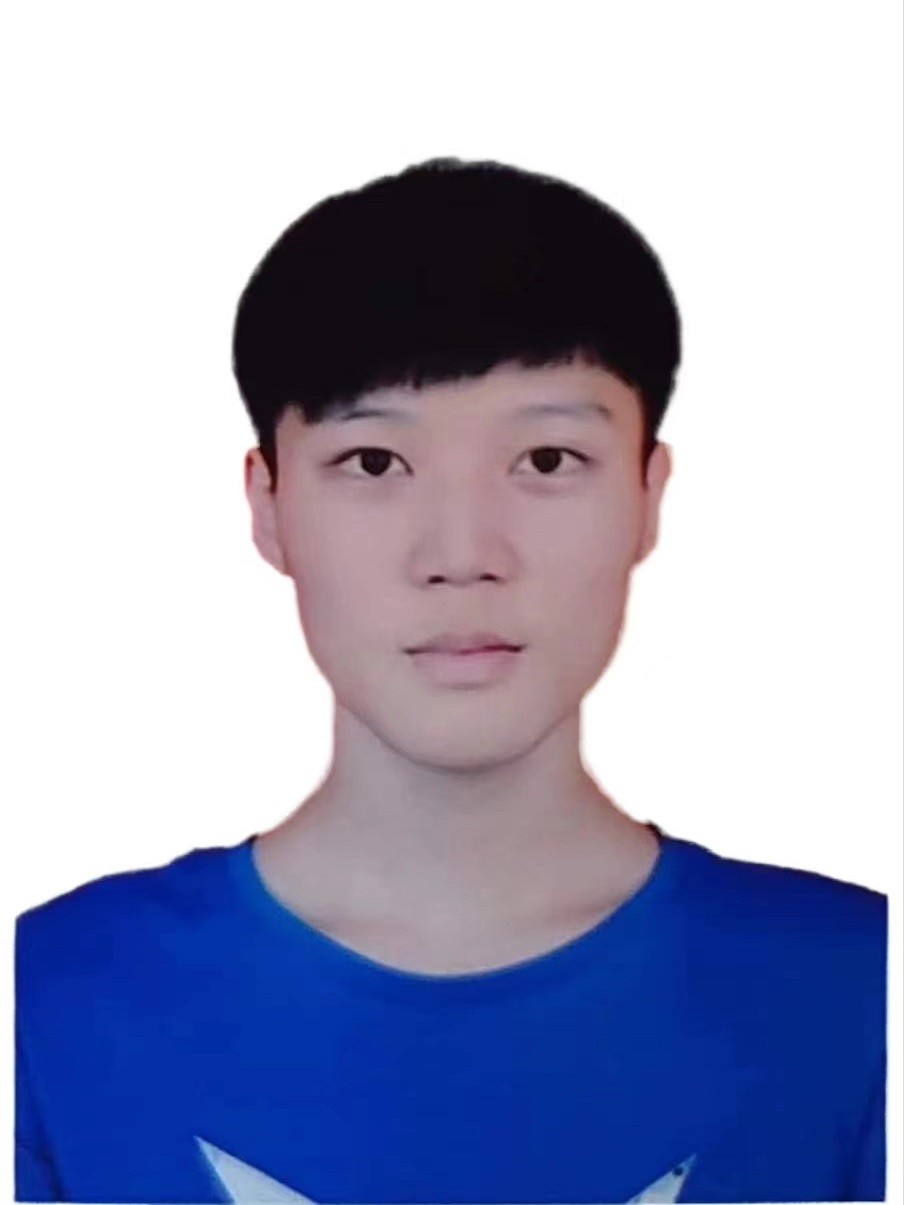}}]{Yupeng Zhu} received the BE degree from Shanghai Jiao Tong University (SJTU), Shanghai, China, in 2024. Since 2024, he has been working toward the master's degree in information and communication engineering at Shanghai Jiao Tong University. His research interests include image representation, image editing and animation.
\end{IEEEbiography}

\begin{IEEEbiography}[{\includegraphics[width=1in,height=1.25in,clip,keepaspectratio]{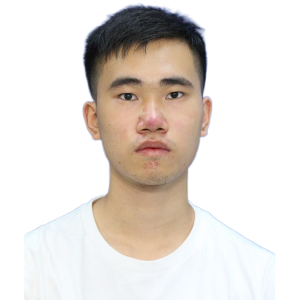}}]{Xiongzhen Zhang} received his Bachelor’s degree from the Macau University of Science and Technology in 2025. He is currently a research intern at Shanghai Jiao Tong University. His research interests include controllable visual generation and physics-based simulation.
\end{IEEEbiography}

\begin{IEEEbiography}
[{\includegraphics[width=1in,height=1.25in,clip,keepaspectratio]{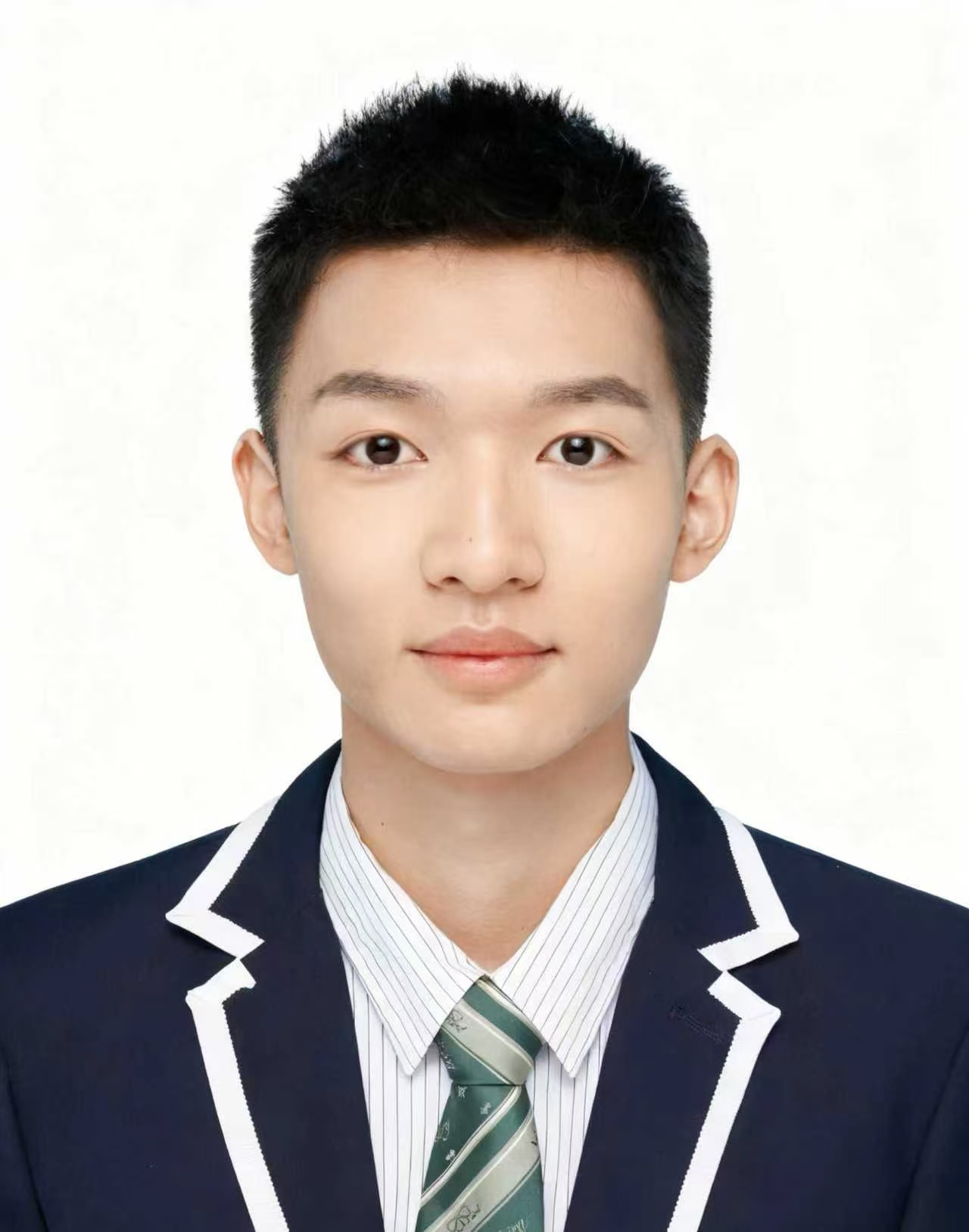}}]
{Zhewen Wan} has been working torwards the BEng degree with the School of Integrated Circuits, Shanghai Jiao Tong University, China, since 2022. His research interests include controllable image generation. 
\end{IEEEbiography}

\begin{IEEEbiography}
[{\includegraphics[width=1in,height=1.25in,clip,keepaspectratio]{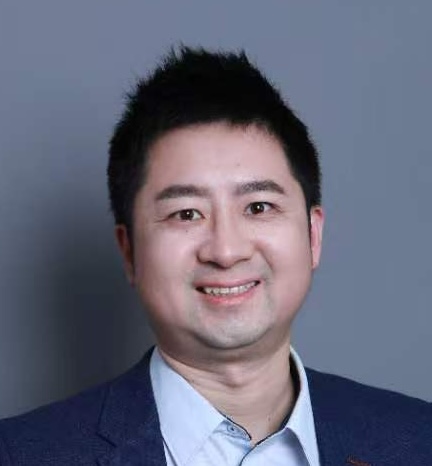}}]
{Yingzhe Li} received his Bachelor’s degree in Electronic and Information Engineering from Beijing University of Posts and Telecommunications in 2006. He then earned his Master’s and Ph.D. degrees from the University of Chinese Academy of Sciences, Shanghai Institute of Microsystem and Information Technology. He is currently working at Huawei, where his work focuses on artificial intelligence, large language models, and the Ascend CANN ecosystem. 
\end{IEEEbiography}

\begin{IEEEbiography}
[{\includegraphics[width=1in,height=1.25in,clip,keepaspectratio]{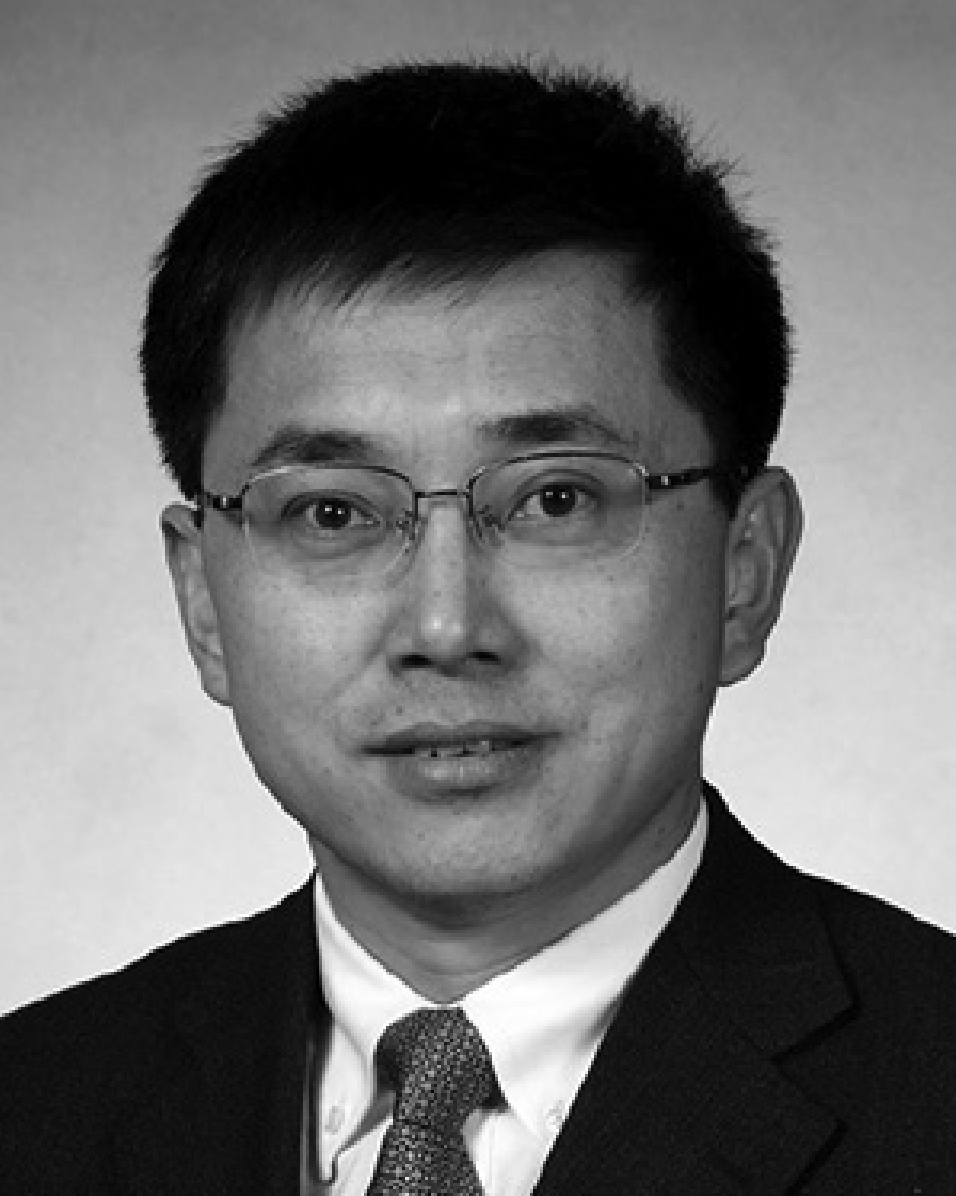}}]
{Wenjun Zhang (Fellow, IEEE)} received the B.S., M.S. and Ph.D. degrees in electronic engineering from Shanghai Jiao Tong University, Shanghai, China, in 1984, 1987 and 1989, respectively. From 1990 to 1993, He worked as a post-doctoral fellow at Philips Kommunikation Industrie AG in Nuremberg, Germany, where he was actively involved in developing HDMAC system. He joined the Faculty of Shanghai Jiao Tong University in 1993 and became a full professor in the Department of Electronic Engineering in 1995. As the national HDTV TEEG project leader, he successfully developed the first Chinese HDTV prototype system in 1998. He was one of the main contributors to the Chinese Digital Television Terrestrial Broadcasting Standard issued in 2006 and is leading team in designing the next generation of broadcast television system in China from 2011. He holds more than 40 patents and published more than 90 papers in international journals and conferences. Prof. Zhangs main research interests include digital video coding and transmission, multimedia semantic processing and intelligent video surveillance. He is a Chief Scientist of the Chinese National Engineering Research Centre of Digital Television (NERCDTV), an industry/government consortium in DTV technology research and standardization and the Chair of Future of Broadcast Television Initiative (FOBTV) Technical Committee.
\end{IEEEbiography}

\begin{IEEEbiography}
[{\includegraphics[width=1in,height=1.25in,clip,keepaspectratio]{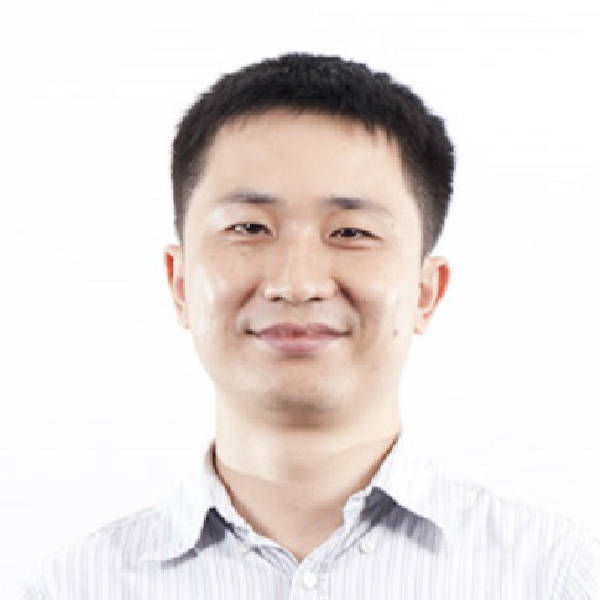}}]
{Bingbing Ni} received the BEng degree in electrical engineering from Shanghai Jiao Tong University, Shanghai, China, in 2005, and the PhD degree from the National University of Singapore, Singapore, in 2011. He is currently a professor with the Department of Electrical Engineering, Shanghai Jiao Tong University. Before that, he was a research scientist with the Advanced Digital Sciences Center, Singapore. He was with Microsoft Research Asia, Beijing, China, as a research intern, in 2009. He was also a software engineer Intern with Google Inc., Mountain View, California, in 2010. He was a recipient of the Best Paper Award from PCM’11 and the Best Student Paper Award from PREMIA’08. He was also the recipient of the first prize in the International Contest on Human Activity Recognition and Localization in conjunction with the International Conference on Pattern Recognition, in 2012.
\end{IEEEbiography}


\vfill

\end{document}